\def\paperID{4774}
\def\confName{CVPR}
\def\confYear{2024}

\def\authorBlock{
    Yang Fu$^1$\footnotemark[1] \qquad
    Sifei Liu$^2$ \qquad
    Amey Kulkarni$^2$\qquad 
    Jan Kautz$^2$ \qquad \\
    Alexei A. Efros$^3$ \qquad 
    Xiaolong Wang$^1$ \\
    $^1$UC San Deigo \qquad 
    $^2$NVIDIA \qquad
    $^3$UC Berkeley \\
}

\newif\ifreview 
\newif\ifarxiv \newcommand{\arxiv}{\arxivtrue}
\newif\ifcamera 
\newif\ifrebuttal 

\arxiv

\pdfoutput=1
\documentclass[10pt,twocolumn,letterpaper]{article}
\ifreview \usepackage[review]{cvpr} \fi
\ifarxiv \usepackage[pagenumbers]{cvpr} \fi
\ifrebuttal \usepackage[rebuttal]{cvpr} \fi
\ifcamera \usepackage{cvpr} \fi

\usepackage{graphicx}	
\usepackage{amsmath}	
\usepackage{amssymb}	
\usepackage{booktabs}
\usepackage{times}
\usepackage{microtype}
\usepackage{epsfig}
\usepackage[table,xcdraw,dvipsnames]{xcolor}
\usepackage{caption}
\usepackage{float}
\usepackage{placeins}
\usepackage{color, colortbl}
\usepackage{stfloats}
\usepackage{enumitem}
\usepackage{tabularx}
\usepackage{xstring}
\usepackage{multirow}
\usepackage{xspace}
\usepackage{url}
\usepackage{subcaption}
\usepackage{xcolor}
\usepackage[hang,flushmargin]{footmisc}
\usepackage{mathtools}
\usepackage{algorithm}
\usepackage{algpseudocode}
\usepackage{tikz}
\usepackage{soul}
\usepackage[percent]{overpic}
\usepackage[accsupp]{axessibility} 

\ifcamera \usepackage[accsupp]{axessibility} \fi

% ...

\ifarxiv  \fi

\newcommand{\customfootnotetext}[2]{{
\renewcommand{\thefootnote}{#1}
\footnotetext[0]{#2}}}

\usepackage{xr-hyper}

\makeatletter
\newcommand*{\addFileDependency}[1]{
  \typeout{(#1)}
  \@addtofilelist{#1}
  \IfFileExists{#1}{}{\typeout{No file #1.}}
}

\makeatother

\definecolor{cvprblue}{rgb}{0.21,0.49,0.74}
\usepackage[pagebackref,breaklinks,colorlinks,citecolor=cvprblue]{hyperref}
\usepackage[capitalize]{cleveref}
\crefname{section}{Sec.}{Secs.}
\crefname{table}{Table}{Tables}
\crefname{figure}{Fig.}{Figs.}

\begin{document}
\title{COLMAP-Free 3D Gaussian Splatting}
\author{\authorBlock}
\twocolumn[{%
\renewcommand\twocolumn[1][]{#1}%
\maketitle
\vspace{-10mm}%
\begin{center}
    \centering
    \includegraphics[width=1.0\textwidth]{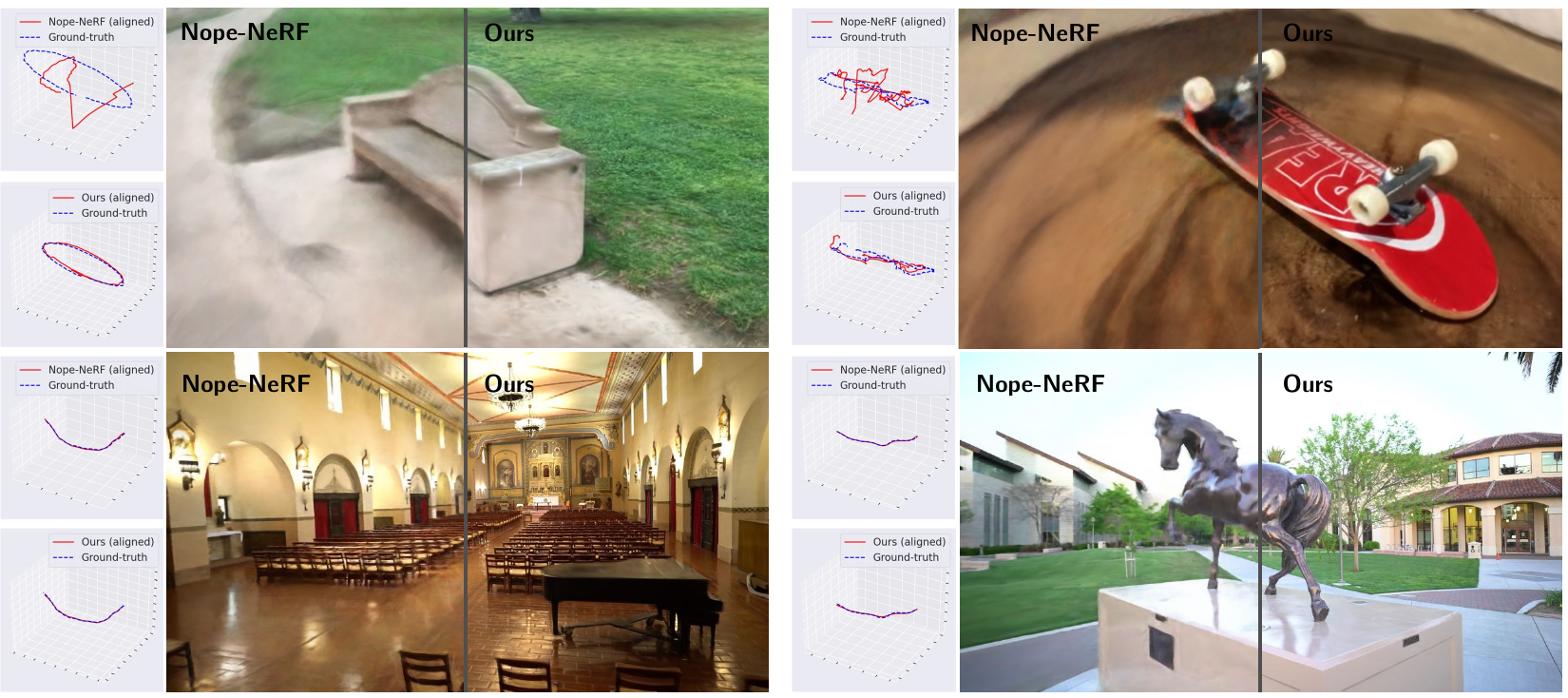}
    \vspace{-5mm}%
    \captionof{figure}{\textbf{Novel View Synthesis and Camera Pose Estimation Comparison.} We propose COLMAP-Free 3D Gaussian Splatting (CF-3DGS) for novel view synthesis without known camera parameters. Our method achieves more robustness in pose estimation and better quality in novel view synthesis than previous state-of-the-art methods.}
    \label{fig:teaser}
\end{center}%
}]

\customfootnotetext{*}{This work was done while Yang Fu was a part-time intern at NVIDIA.}

\begin{abstract}
While neural rendering has led to impressive advances in scene reconstruction and novel view synthesis, it relies heavily on accurately pre-computed camera poses. To relax this constraint, multiple efforts have been made to train Neural Radiance Fields (NeRFs) without pre-processed camera poses. However, the implicit representations of NeRFs provide extra challenges to optimize the 3D structure and camera poses at the same time. On the other hand, the recently proposed 3D Gaussian Splatting provides new opportunities given its explicit point cloud representations. This paper leverages both the explicit geometric representation and the continuity of the input video stream to perform novel view synthesis without any SfM preprocessing. We process the input frames in a sequential manner and progressively grow the 3D Gaussians set by taking one input frame at a time, without the need to pre-compute the camera poses. Our method significantly improves over previous approaches in view synthesis and camera pose estimation under large motion changes. Our project page is \url{https://oasisyang.github.io/colmap-free-3dgs}.
\end{abstract}

\section{Introduction}
\label{sec:intro}

The field of photo-realistic scene reconstruction and view synthesis has been largely advanced with the rise of Neural Radiance Fields (NeRFs~\cite{mildenhall2021nerf}). An important initialization step for training NeRF is to first prepare the camera poses for each input image. This is usually achieved by running the Structure-from-Motion (SfM) library COLMAP~\cite{schonberger2016structure}. However, this pre-processing is not only time-consuming but also can fail due to its sensitivity to feature extraction errors and difficulties in handling textureless or repetitive regions. 

Recent studies~\cite{wang2021nerfmm, lin2021barf, bian2023nope} have focused on reducing the reliance on SfM by integrating pose estimation directly within the NeRF framework. Simultaneously solving 3D scene reconstruction and camera registration has been a chicken-and-egg problem for a long time in computer vision. 
This challenge is further amplified in the context of NeRF and its implicit representation, where the optimization process often involves additional constraints.
For instance, BARF~\cite{lin2021barf} requires initial poses that are close to their ground truth locations,
and NeRFmm~\cite{wang2021nerfmm} is largely limited to face-forwarding scenes. The recently proposed Nope-NeRF~\cite{bian2023nope} takes a long time to train (30 hours) and does not work well when the camera pose changes a lot (e.g., 360 degrees), as shown in the two top cases in Fig.~\ref{fig:teaser}. Fundamentally, NeRFs optimize camera parameters in an indirect way, by updating the ray casting from camera positions, which makes optimization challenging.

The arrival of 3D Gaussian Splatting~\cite{kerbl20233d} extends the volumetric rendering in NeRFs to accommodate point clouds. While it was originally proposed with pre-computed cameras, we find it offers a new opportunity to perform view synthesis without SfM pre-processing. We propose \textbf{C}OLMAP-\textbf{F}ree \textbf{3D} \textbf{G}aussian \textbf{S}platting (CF-3DGS), which leverages two key ingredients: the \textbf{temporal continuity from video} and the \textbf{explicit point cloud representation}. 
We summarize out approach below.

Instead of optimizing with all the frames at once, we propose to build the 3D Gaussians of the scene in a continuous manner, ``growing'' one frame at a time as the camera moves. In this process, we will extract a \emph{local} 3D Gaussians set for each frame, and also maintain a \emph{global} 3D Gaussians set of the whole scene. Assuming we are iterating through $t=\{1,...,T\}$ frames in a sequential manner, we perform a two-step procedure each time: (i) We construct a local 3D Gaussians set given frame $t-1$, and we sample the next nearby frame $t$. Our goal is to learn an affine transformation that can transform the 3D Gaussians in frame $t-1$ to render the pixels in frame $t$. Neural rendering provides the gradients for optimizing the affine transformation, which is essentially the relative camera pose between frames $t-1$ and $t$. This optimization is not difficult as the \textbf{explicit} point cloud representation allows us to directly apply an affine transformation on it which cannot be achieved with NeRFs, and the two frames are close (\textbf{temporal continuity}) which makes the transformation relatively small. (ii) Once we have the relative camera pose between frames $t-1$ and $t$, we can infer the relative pose between the first frame and frame $t$. This allows us to aggregate the current frame information into the global 3D Gaussians set, where we will perform optimization with the current and all the previous frames and camera poses. 

We experiment with two datasets: the Tanks and Temples dataset~\cite{Knapitsch2017} and videos randomly selected from the CO3D dataset~\cite{reizenstein2021common}. We evaluate both view synthesis and camera pose estimation tasks, and compare with previous approaches without pre-computed camera poses. Our method performs significantly better than previous approaches on view synthesis in both datasets. For camera pose estimation, we find our method performs on par with the most recent Nope-NeRF~\cite{bian2023nope} when the camera motion is small, and outperforms all approaches by a large margin when the camera changes a lot, such as in the 360-degree videos in CO3D.

\section{Related Work}
\label{sec:related}

\noindent\textbf{Novel View Synthesis.}
To generate realistic images from novel viewpoints, several different 3D scene representations have been employed, such as planes~\cite{horry1997tour,hoiem2005automatic}, meshes~\cite{hu2020worldsheet,Riegler2020FVS, riegler2021stable}, point clouds~\cite{xu2022point,zhang2022differentiable}, and multi-plane images~\cite{tucker2020single, zhou2018stereo, li2021mine}. Recently, NeRFs~\cite{mildenhall2021nerf} have gained prominence in this field due to its exceptional capability of photorealistic rendering. Several efforts have been made on top of the vanilla NeRF for advanced rendering quality. These improvements include enhancing anti-aliasing effects~\cite{barron2021mip, barron2022mip, barron2023zip, zhang2020nerf++}, refining reflectance~\cite{verbin2022ref, Attal_2023_CVPR}, training with sparse view~\cite{kim2022infonerf, niemeyer2022regnerf, xu2022sinnerf, yang2023freenerf, irshad2023neo},   and reducing training times~\cite{yu_and_fridovichkeil2021plenoxels, mueller2022instant, reiser2021kilonerf} and rendering time~\cite{liu2020neural, garbin2021fastnerf, sun2022direct,yu2021plenoctrees}. 

More recently, point-cloud-based representation~\cite{xu2022point, zhang2022differentiable, kerbl20233d, luiten2023dynamic, kopanas2022neural, yifan2019differentiable} has been widely used for its efficiency during rendering. For instance, Zhang~\etal~\cite{zhang2022differentiable} propose to learn the per-point position and view-dependent appearance, using a differentiable splat-based renderer, from point clouds initialized from object masks. Additionally, 3DGS~\cite{kerbl20233d} enables real-time rendering of novel views by its pure explicit representation and the novel differential point-based splatting method.
However, most of these approaches still rely on pre-computed camera parameters obtained from SfM algorithms~\cite{hartley2003multiple, schonberger2016structure, mur2015orb, taketomi2017visual}.

\noindent\textbf{NeRF without SfM Preprocessing.} 
Recently, there has been growing interest 
in trying to eliminate the required preprocessing step of camera estimation in NeRFs.
The initial effort in this direction was i-NeRF~\cite{yen2021inerf}, which predicts camera poses by matching keypoints using a pre-trained NeRF. Following this, NeRFmm~\cite{wang2021nerfmm} introduced a method to jointly optimize the NeRF network and camera pose embeddings. However, despite its successor, SiNeRF~\cite{xia2022sinerf}, employing SIREN-MLP~\cite{sitzmann2020implicit} and a mixed region sampling strategy to address NeRFmm's sub-optimal issues, it remains limited to forward-facing scenes. BARF~\cite{lin2021barf} and GARF~\cite{chng2022garf} propose to alleviate the gradient inconsistency issue caused by high-frequency parts of positional embedding. For instance, BARF proposes a coarse-to-fine positional encoding strategy for camera poses and NeRF joint optimization. Though they could handle more complex camera motions, they require a good initialization from the ground-truth cameras.(\eg, within 15$^\circ$ of the ground-truth). More advanced works~\cite{bian2023nope,meuleman2023progressively,cheng2023lu} seek help from some pre-trained networks, \ie, monocular depth estimation and optical flow estimation, to obtain prior knowledge of geometry or correspondence. For example, Nope-NeRF~\cite{bian2023nope} trains a NeRF by incorporating undistorted depth priors which are correct from the monocular depth estimation during training. Additionally, VideoAE~\cite{Lai21a}, RUST~\cite{sajjadi2023rust}, MonoNeRF~\cite{fu2023mononerf} and FlowCam~\cite{smith2023flowcam} learn a generalizable scene representation from unposed videos, but their view synthesis performance is unsatisfactory without per-scene optimization.

In summary, although showing some promising results, prior works on NeRFs with unknown poses assume either small perturbations~\cite{lin2021barf, chng2022garf}, a narrow range of camera motion~\cite{xia2022sinerf, wang2021nerfmm}, or prior knowledges~\cite{bian2023nope, meuleman2023progressively}. These approaches face difficulties when handling challenging camera trajectories with large camera motion, \eg, 360$^\circ$ scenes in the CO3D~\cite{reizenstein2021common} dataset. Furthermore, most existing works require quite a long training time, typically exceeding 10 hours. To overcome these limitations, our work proposes a joint optimization of camera parameters and 3D Gaussians, utilizing both local and global 3DGS strategies.

\begin{figure*}[tp]
    \centering
    \includegraphics[width=1.0\textwidth]{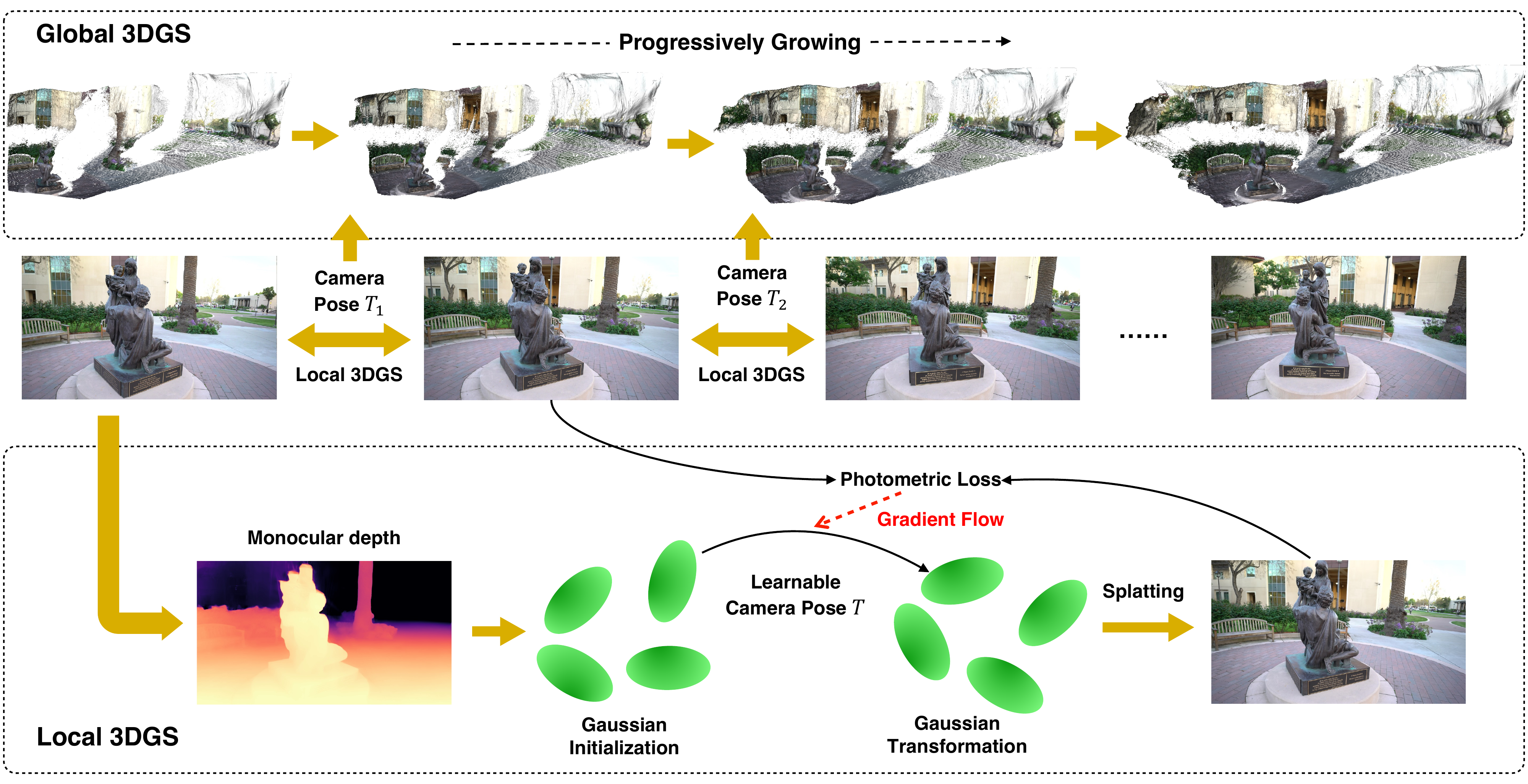}
    \vspace{-4mm}
    \caption{\textbf{Overview of proposed CF-3DGS}. Our method takes a sequence of images as input to learn a set of 3D Gaussian that presents the input scene and jointly estimates the camera poses of the frames. We first introduce a local 3DGS to estimate the relative pose of two nearby frames by approximating the Gaussian transformation. Then, a global 3DGS is utilized to model the scene by progressively growing the set of 3D Gaussian as the camera moves.}
    \label{fig:method}
    \vspace{-4mm}
\end{figure*}
\section{Method}
\label{sec:method}

Given a sequence of unposed images along with camera intrinsics, our goal is to recover the camera poses and reconstruct the photo-realistic scene. To this end, we propose CF-3DGS to optimize the 3D Gaussian Splatting (3DGS~\cite{kerbl20233d}) and camera poses simultaneously. We detail our method in the following sections, starting from a brief review of the representation and rendering process of 3DGS in Sec.~\ref{method:3dgs}. Then, we propose a \textit{local} 3DGS, a simple yet effective method to estimate the relative camera pose from each pair of nearby frames, in Sec.~\ref{method:pose}. Finally, we introduce a \textit{global} 3DGS, featuring a progressive expansion of the 3D Gaussians from unobserved views in sequential order, in Sec.~\ref{method:joint}.

\subsection{Preliminary: 3D Gaussian Splatting}\label{method:3dgs}

3DGS~\cite{kerbl20233d} models the scene as a set of 3D Gaussians, which is an explicit form of representation, in contrast to the implicit representation used in NeRFs. Each Gaussian is characterized by a covariance matrix $\Sigma$ and a center (mean) point~$\mu$, 
\begin{equation}
    G(x) = e^{-\frac{1}{2}(x-\mu)^{\top}\Sigma^{-1}(x-\mu)}
    \label{eq:gauss}
\end{equation}
The means of 3D Gaussians are initialized by a set of sparse point clouds(\eg, always obtained from SfM). 
Each Gaussian is parameterized as the following parameters: (a) a center position $\mu \in \mathbb{R}^3$; (b) spherical harmonics (SH) coefficients $c \in \mathbb{R}^k$ (k represents the degrees of freedom) that represents the color; (c) rotation factor $r \in \mathbb{R}^4$ (in quaternion rotation); (d) scale factor $s \in \mathbb{R}^3$; (e) opacity $\alpha \in \mathbb{R}$.
Then, the covariance matrix $\Sigma$ describes an ellipsoid configured by a scaling matrix $S=\text{diag}([s_x, s_y, s_z])$ and rotation matrix $R=\text{q2R}([r_w, r_x, r_y, r_z])$, where $q2R()$ is the formula for constructing a rotation matrix from a quaternion. Then, the covariance matrix can be computed as follows,
\begin{equation}
    \Sigma = R S S^{\top} R^{\top}
    \label{eq:cov}
\end{equation}

In order to optimize the parameters of 3D Gaussians to represent the scene, we need to render them into images in a differentiable manner. As introduced in~\cite{kerbl20233d}, the rendering from a given camera view $W$ involves the process of splatting the Gaussian onto the image plane, which is achieved by approximating the projection of a 3D Gaussian along the depth dimension into pixel coordinates. Given a viewing transform $W$ (also known as the camera pose), the covariance matrix $\Sigma^{\text{2D}}$ in camera coordinates can be expressed by
\begin{equation}
    \Sigma^{\text{2D}} = J W \Sigma W^{\top} J^{\top}
    \label{eq:cov_cam}
\end{equation}
where $J$ is the Jacobian of the affine approximation of the projective transformation. For each pixel, the color and opacity of all the Gaussians are computed using Eq.~\ref{eq:gauss}, and the final rendered color can be formulated as the alpha-blending of $N$ ordered points that overlap the pixel, 
\begin{equation}
    C_{pix} = \sum_{i}^{N} c_i \alpha_i \prod_j^{i-1}(1-\alpha_j)
    \label{eq:render}
\end{equation}
where $c_i, \alpha_i$ represents the density and color of this point computed from the learnable per-point opacity and SH color coefficients weighted by the Gaussian covariance $\Sigma$, which we ignore in Eq.~\ref{eq:render} for simplicity. 

To perform scene reconstruction, given the ground truth poses that determine the projections, we fit a set of initialized Gaussian points to the desired objects or scenes by learning their parameters, i.e., $\mu$ and $\Sigma$. With the differentiable renderer as in Eq.~\ref{eq:render}, all those parameters, along with the SH and opacity, can be easily optimized through a photometric loss. In our approach, we reconstruct scenes following the same process, but replacing the ground truth poses with the estimated ones, as detailed in the following sections.

\subsection{Local 3DGS for Relative Pose Estimation}\label{method:pose}

Previous studies~\cite{lin2021barf, jeong2021self, bian2023nope} have demonstrated the feasibility of simultaneously estimating camera parameters and optimizing a Neural Radiance Field (NeRF). This typically involves the integration of various regularization terms and geometric priors. However, rather than directly optimizing camera poses, most existing methods prioritize optimizing the ray casting process from varying camera positions. This is dictated by the nature of implicit representation and the implementation of ray tracing in NeRFs. This indirect approach often results in a complex and challenging optimization under large camera movement scenarios.

On the other hand, 3DGS~\cite{kerbl20233d} utilizes an explicit scene representation in the form of point clouds enabling straightforward deformation and movement, as demonstrated in its recent application to dynamic scenes~\cite{luiten2023dynamic, wu20234d}. To take advantage of 3DGS, we introduce a \textit{local} 3DGS to estimate the relative camera pose.

We reveal the relationship between the camera pose and the 3D rigid transformation of Gaussian points, in the following. Given a set of 3D Gaussians with centers $\mu$, projecting them with the camera pose $W$ yields:
\begin{equation}
    \centering
    \mu_{\text{2D}} = K (W \mu) / (W \mu)_{z}
\end{equation}
where $K$ is the intrinsic projection matrix. Alternatively, the 2D projection $\mu_{\text{2D}}$ can be obtained from the orthogonal direction $\mathbb{I}$ of a set of rigidly transformed points, i.e., $\mu^{\prime} = W \mu$, which yields $\mu_{\text{2D}} \coloneqq K (\mathbb{I} \mu^{\prime}) / (\mathbb{I} \mu^{\prime})_{z}$. As such, estimating the camera poses $W$ is equivalent to estimating the transformation of a set of 3D Gaussian points. Based on this finding, we designed the following algorithm to estimate the relative camera pose.

\noindent\textbf{Initialization from a single view.} 
As demonstrated in Fig.~\ref{fig:method} (bottom part), given a frame $I_t$ at timestep $t$, we first utilize an off-the-shelf monocular depth network, \ie, DPT~\cite{ranftl2021vision}, to generate the monocular depth, denoted as $D_t$. Given that monocular depth $D_t$ offers strong geometric cues without needing camera parameters, we initialize 3DGS with points lifted from monocular depth, leveraging camera intrinsic and identity camera pose, instead of the original SfM points. After initialization, we learn a set of 3D Gaussian $G_t$ with all attributes to minimize the photometric loss between the rendered image and the current frame $I_t$, 
\begin{equation}
    {G_t}^* = \arg \min_{c_t, r_t, s_t, \alpha_t} \mathcal{L}_{rgb}(\mathcal{R}(G_t), I_t),
\end{equation}
where $\mathcal{R}$ is the 3DGS rendering process. The photometric loss  $\mathcal{L}_{rgb}$ is $\mathcal{L}_1$ combined with a D-SSIM:
\begin{equation}
    \mathcal{L}_{rgb} = (1-\lambda) \mathcal{L}_1 + \lambda \mathcal{L}_{\text{D-SSIM}}
\end{equation}
We use $\lambda=0.2$ for all experiments. This step is quite lightweight to run and only takes around 5s to fit the input frame $I_t$. 

\noindent\textbf{Pose Estimation by 3D Gaussian Transformation.} 
To estimate the relative camera pose, we transform the pre-trained 3D Gaussian ${G_t}^*$ by a learnable SE-3 affine transformation $T_t$ into frame $t+1$, denoted as $G_{t+1}= T_t \odot G_t$. The transformation $T_t$ is optimized by minimizing the photometric loss between the rendered image and the next frame $I_{t+1}$
\begin{equation}
    {T_t}^* = \arg \min_{T_t} \mathcal{L}_{rgb}(\mathcal{R}(T_t \odot G_t), I_{t+1}),
\end{equation}
Note that during the aforementioned optimization process, we freeze all attributes of the pre-trained 3D Gaussian ${G_t}^*$ to separate the camera movement from the deformation, densification, pruning, and self-rotation of the 3D Gaussian points. 
The transformation $T$ is represented in form of quaternion rotation $\mathbf{q} \in \mathfrak{so}(3)$ and translation vector $\mathbf{t} \in \mathbb{R}^3$. As two adjunct frames are close, the transformation is relatively small and easier to optimize. Similar to the initialization phase, the pose optimization step is also quite efficient, typically requiring only 5-10 seconds.

\subsection{Global 3DGS with Progressively Growing}\label{method:joint}
By employing the local 3DGS on every pair of images, we can infer the relative pose between the first frame and any frame at timestep $t$. However, these relative poses could be noisy resulting in a dramatic impact on optimizating a 3DGS for the whole scene (see Table~\ref{table:abs_grow}). To tackle this issue, we propose to learn a global 3DGS progressively in a sequential manner.

As described in the top part of Fig.~\ref{fig:method}, starting from the $t$th frame $I_t$, we first initialize a set of 3D Gaussian points with the camera pose set as orthogonal, as aforementioned. Then, utilizing the local 3DGS, we estimate the relative camera pose between frames $I_t$ and $I_{t+1}$. Following this, the global 3DGS updates the set of 3D Gaussian points, along with all attributes, over $N$ iterations, using the estimated relative pose and the two observed frames as inputs. As the next frame $I_{t+2}$ becomes available, this process is repeated: we estimate the relative pose between $I_{t+1}$ and $I_{t+2}$, and subsequently infer the relative pose between $I_{t}$ and $I_{t+2}$. 

To update the global 3DGS to cover the new view, we densify the Gassians that are "under-reconstruction" as new frames arrive. As suggested in~\cite{kerbl20233d}, we determine the candidates for densification by the average magnitude of view-space position gradients. Intuitively, the unobserved frames always contain regions that are not yet well reconstructed, and the optimization tries to move the Gaussians to correct with a large gradient step. Therefore, to make the densification concentrate on the unobserved content/regions, we densify the global 3DGS every $N$ steps that aligns with the pace of adding new frames. In addition, instead of stopping the densification in the middle of the training stage, we keep growing the 3D Gaussian points until the end of the input sequence.
By iteratively applying both local and global 3DGS, the global 3DGS will grow progressively from the initial partial point cloud to the completed point cloud that covers the whole scene throughout the entire sequence, and simultaneously accomplish photo-realistic reconstruction and accurate camera pose estimation.

\begin{table*}[ht]
\setlength{\tabcolsep}{3.5pt}
\centering
\footnotesize
\begin{tabular}{cccccccccccccccccccccc}
\hline
& \multirow{2}{*}{scenes} &  & \multicolumn{3}{c}{Ours} &  & \multicolumn{3}{c}{Nope-NeRF}  &  & \multicolumn{3}{c}{BARF} &  & \multicolumn{3}{c}{NeRFmm} &  & \multicolumn{3}{c}{SC-NeRF} \\ 
\cline{4-6} \cline{8-10} \cline{12-14} \cline{16-18} \cline{20-22} &  &  & PSNR $\uparrow$  & SSIM $\uparrow$  & LPIPS $\downarrow$   &  & PSNR  & SSIM  & LPIPS &  & PSNR & SSIM  & LPIPS  &  & PSNR   & SSIM & LPIPS  &  & PSNR  & SSIM  & LPIPS  \\ \hline
  & Church                  &  & \textbf{30.23} & \textbf{0.93} & \textbf{0.11}  & & 25.17 & 0.73 & 0.39 &  & 23.17 & 0.62          & 0.52  &  & 21.64   & 0.58    & 0.54   &  & 21.96 & 0.60 & 0.53          \\
  & Barn                    &  & \textbf{31.23} & \textbf{0.90} & \textbf{0.10}  & & 26.35 & 0.69 & 0.44 &  & 25.28 & 0.64          & 0.48  &  & 23.21   & 0.61    & 0.53   &  & 23.26 & 0.62 & 0.51          \\
  & Museum                  &  & \textbf{29.91} & \textbf{0.91} & \textbf{0.11} & & 26.77 & 0.76 & 0.35 &  & 23.58 & 0.61          & 0.55  &  & 22.37   & 0.61    & 0.53   &  & 24.94 & 0.69 & 0.45          \\
  & Family                  &  & \textbf{31.27} & \textbf{0.94} & \textbf{0.07} & & 26.01 & 0.74 & 0.41 &  & 23.04 & 0.61          & 0.56  &  & 23.04   & 0.58    & 0.56   &  & 22.60 & 0.63 & 0.51          \\
  & Horse                   &  & \textbf{33.94} & \textbf{0.96} & \textbf{0.05} & & 27.64 & 0.84 & 0.26 &  & 24.09 & 0.72          & 0.41  &  & 23.12   & 0.70    & 0.43   &  & 25.23 & 0.76 & 0.37          \\
  & Ballroom                &  & \textbf{32.47} & \textbf{0.96} & \textbf{0.07} & & 25.33 & 0.72 & 0.38 &  & 20.66 & 0.50          & 0.60  &  & 20.03   & 0.48    & 0.57   &  & 22.64 & 0.61 & 0.48          \\
  & Francis                 &  & \textbf{32.72} & \textbf{0.91} & \textbf{0.14} & & 29.48 & 0.80 & 0.38 &  & 25.85 & 0.69          & 0.57  &  & 25.40   & 00.69   & 0.52   &  & 26.46 & 0.73 & 0.49          \\
  & Ignatius                &  & \textbf{28.43} & \textbf{0.90} & \textbf{0.09} & & 23.96 & 0.61 & 0.47 &  & 21.78 & 0.47          & 0.60  &  & 21.16   & 0.45    & 0.60   &  & 23.00 & 0.55 & 0.53          \\ \hline
  & mean                    &  & \textbf{31.28} & \textbf{0.93} & \textbf{0.09} & & 26.34 & 0.74 & 0.39 &  & 23.42 & 0.61          & 0.54  &  & 22.50   & 0.59    & 0.54   &  & 23.76 & 0.65 & 0.48          \\ \hline
\end{tabular}
\caption{\textbf{Novel view synthesis results on Tanks and Temples}. Each baseline method is trained with its public code under the original settings and evaluated with the same evaluation protocol. The best results are highlighted in bold.}
\label{table:nvs}
\end{table*}

\section{Experiments}
\label{sec:exp}

\subsection{Experimental Setup}
\noindent\textbf{Datasets.}
We conduct extensive experiments on different real-world datasets, including Tanks and Temples~\cite{Knapitsch2017} and CO3D-V2~\cite{reizenstein2021common}. \textbf{Tanks and Temples:} Similar to~\cite{bian2023nope}, we evaluate novel view synthesis quality and pose estimation accuracy on 8 scenes covering both indoor and outdoor scenes. For each scene, we sample 7 images from every 8-frame clip as training samples and test the novel view synthesis quality on the remaining 1/8 images. The camera poses are estimated and evaluated on all training samples after Umeyama alignment~\cite{umeyama1991least}. \textbf{CO3D-V2:} It consists of thousands of object-centric videos where the whole object is kept in view while moving a full circle around it. Compared with Tanks and Temples, recovering camera poses from CO3D videos is much harder as it involves large and complicated camera motions. We randomly select 5 scenes\footnote{We specify all selected scenes in the supplementary material.} of different categories objects and follow the same protocol to split the train/test set.

\noindent\textbf{Metrics.}
We evaluate the tasks of novel view synthesis and camera pose estimation. For novel view synthesis, we 
For camera pose estimation, we report the camera rotation and translation error, including the Absolute Trajectory Error (ATE) and Relative Pose Error (RPE) as in~\cite{lin2021barf, bian2023nope}. For novel view synthesis, we report the standard evaluation metrics including PSNR, SSIM~\cite{wang2004image}, and LPIPS~\cite{zhang2018unreasonable}.

\noindent\textbf{Implementation Details.} Our implementation is primarily based on the PyTorch~\cite{paszke2017automatic} framework and we follow the optimization parameters by the configuration outlined in the 3DGS~\cite{kerbl20233d} unless otherwise specified. Notably, we set the number of steps of adding new frames equal to the intervals of point densification, in order to achieve progressive growth of the whole scene. Further, we keep resetting the opacity until the end of the training process, which enables us to integrate the new frames into the Gaussian model established from observed frames. Moreover, the camera poses are optimized in the representation of quaternion rotation  $\mathbf{q} \in \mathfrak{so}(3)$ and translation vector $\mathbf{t} \in \mathbb{R}^3$. The initial learning rate is set to $10^{-5}$ and gradually decay to $10^{-6}$ until convergence. All experiments are conducted on a single single RTX 3090 GPU. More details are provided in the supplementary material.

\subsection{Comparing with Pose-Unknown Methods}
In this subsection, we compare our method with several baselines including, Nope-NeRF~\cite{bian2023nope}, BARF~\cite{lin2021barf}, NeRFmm~\cite{wang2021nerfmm} and SC-NeRF~\cite{jeong2021self} on both novel view synthesis and camera pose estimation. 
\begin{table*}[t]
\setlength{\tabcolsep}{4pt}
\centering
\footnotesize
\begin{tabular}{cccccccccccccccccccccc} 
\hline
\multirow{2}{*}{} & \multirow{2}{*}{scenes} &  & \multicolumn{3}{c}{Ours} &  & \multicolumn{3}{c}{Nope-NeRF} &  &  \multicolumn{3}{c}{BARF} &  & \multicolumn{3}{c}{NeRFmm}  &  & \multicolumn{3}{c}{SC-NeRF}   \\ 
\cline{4-6} \cline{8-10} \cline{12-14} \cline{16-18} \cline{20-22} &  &  & $\text{RPE}_t \downarrow$ & $\text{RPE}_r \downarrow$ & ATE$ \downarrow$ &  & $\text{RPE}_t $ & $\text{RPE}_r $ & ATE &  &  $\text{RPE}_t $ & $\text{RPE}_r $ & ATE &  & $\text{RPE}_t$ & $\text{RPE}_r$ & ATE   &  & $\text{RPE}_t$ & $\text{RPE}_r$ & ATE    \\ 
\hline

& Church                  &  & 0.008 & 0.018 & 0.002 &  & 0.034 & 0.008  & 0.008  &  & 0.114 &0.038 & 0.052 &  & 0.626          & 0.127          & 0.065 &  & 0.836          & 0.187          & 0.108  \\
& Barn                    &  & 0.034 & 0.034 & 0.003  &  & 0.046 & 0.032  &  0.004 &  & 0.314 &0.265 & 0.050  &  & 1.629          & 0.494          & 0.159 &  & 1.317          & 0.429          & 0.157  \\
& Museum                  &  & 0.052 & 0.215 & 0.005  &  &  0.207 & 0.202 & 0.020   &  & 3.442 &1.128 & 0.263   &  & 4.134          & 1.051          & 0.346 &  & 8.339          & 1.491          & 0.316  \\
& Family                  &  & 0.022 & 0.024 & 0.002 & & 0.047 & 0.015 & 0.001  &  & 1.371 &0.591 & 0.115  &  & 2.743 & 0.537 & 0.120 &  & 1.171 & 0.499 & 0.142  \\
& Horse                   &  & 0.112 & 0.057 & 0.003 & & 0.179 & 0.017 & 0.003   &  & 1.333 &0.394 & 0.014  &  & 1.349          & 0.434          & 0.018 &  & 1.366          & 0.438          & 0.019  \\
& Ballroom                &  & 0.037 & 0.024 & 0.003 & & 0.041  & 0.018 & 0.002   &  & 0.531 &0.228 & 0.018  &  & 0.449          & 0.177          & 0.031 &  & 0.328          & 0.146          & 0.012  \\
& Francis                 &  & 0.029 & 0.154 & 0.006 &  &  0.057 & 0.009  & 0.005   &  & 1.321 &0.558 & 0.082  &  & 1.647          & 0.618          & 0.207 &  & 1.233          & 0.483          & 0.192  \\
& Ignatius                &  & 0.033 & 0.032 & 0.005 & &  0.026  & 0.005  & 0.002   &  & 0.736 &0.324 & 0.029  &  & 1.302          & 0.379          & 0.041 &  & 0.533          & 0.240          & 0.085  \\ \hline
& mean                    &  & \textbf{0.041} & 0.069 & \textbf{0.004} & & 0.080  & \textbf{0.038} & 0.006  &  & 1.046 &0.441 & 0.078   &  & 1.735 &0.477 & 0.123 &  & 1.890 &0.489 & 0.129   \\

\hline
\end{tabular}
\caption{\textbf{Pose accuracy on Tanks and Temples}. 
Note that we use COLMAP poses in Tanks and Temples as the ``ground truth". The unit of $\text{RPE}_r$ is in degrees, ATE is in the ground truth scale and $\text{RPE}_t$ is scaled by 100. The best results are highlighted in bold.
}
\vspace{-4mm}
\label{table:pose}
\end{table*}
\begin{figure*}[tp]
    \centering
    \includegraphics[width=1.0\linewidth]{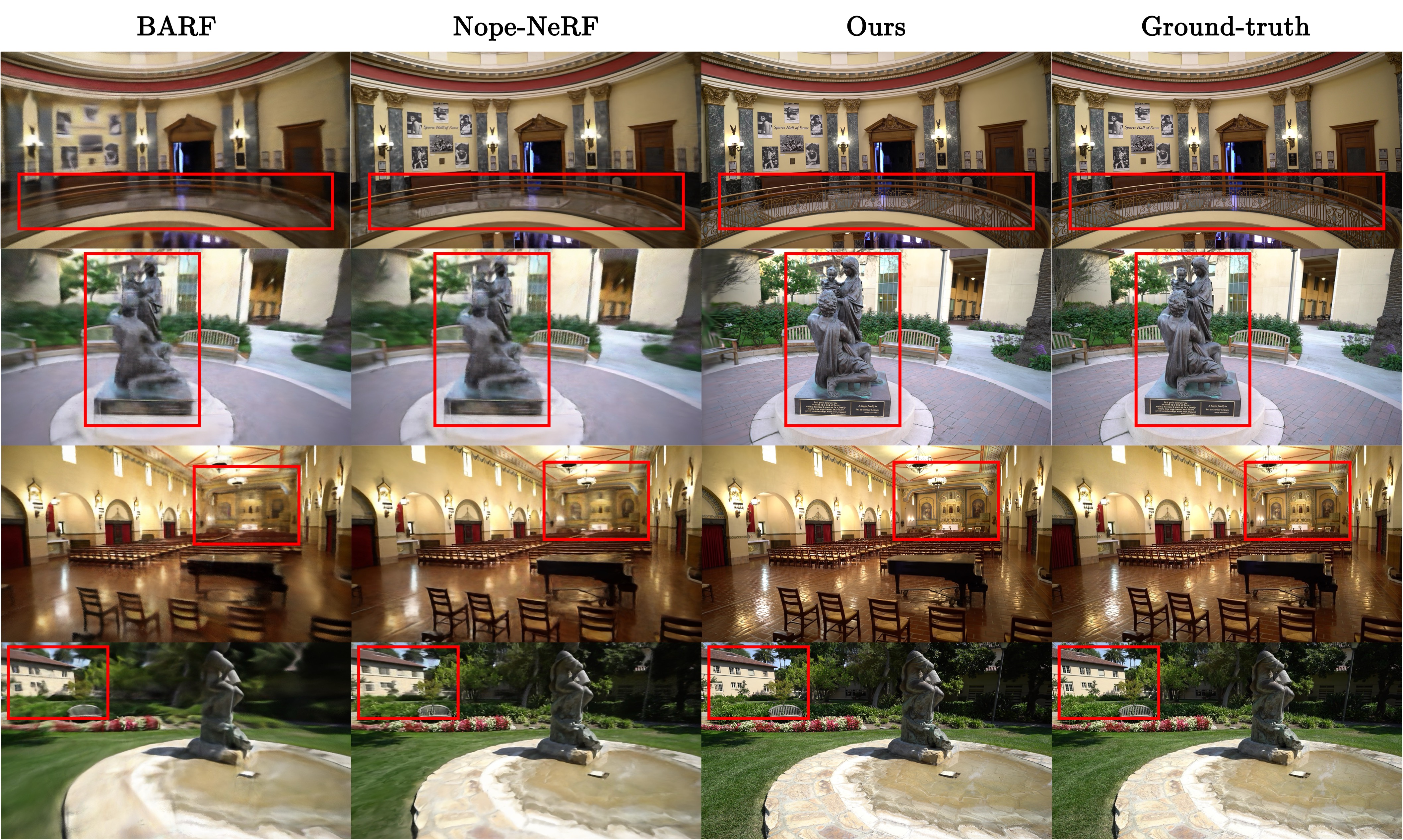}
    \caption{\textbf{Qualitative comparison for novel view synthesis on Tanks and Temples.} Our approach produces more realistic rendering results than other baselines. \textbf{Better viewed when zoomed in.}}
    \label{fig:nvs}
\end{figure*}

\textbf{Novel View Synthesis.} Different from the standard setting where the camera poses of testing views are given, we need to first obtain the camera poses of test views for rendering. Inspired by NeRFmm~\cite{wang2021nerfmm}, we freeze the pre-trained 3DGS model that trained on the training views, while optimizing testing views' camera poses via minimizing the photometric error between the synthesised images and the test views. 
To speed up convergence, the camera pose of each test view is initialised by the closest camera position from the learnt camera poses of all training frames, which are then fine-tuned with the photometric error to obtain the testing camera pose. The same procedure is performed on all baselines for a fair comparison. 

We report the comparison results on Tanks and Temples in Table~\ref{table:nvs}. Our method consistently consistently outperforms the others across all metrics. Notably, compared to Nope-NeRF~\cite{bian2023nope} which takes a significantly longer training time, our approach achieves superior results within a shorter training duration (\eg, 25 hrs vs.\ 1.5 hrs).
We also show the qualitative results in Fig.~\ref{fig:nvs}. As illustrated in Fig.~\ref{fig:nvs}, The images synthesized through our approach are significantly sharper and clearer than those produced by the other methods, as evidenced by the notably higher scores in terms of SSIM and LPIPS, as shown in Table \ref{table:nvs}.
\begin{table*}[ht]
\centering
\setlength{\tabcolsep}{2.5pt}
\footnotesize
\begin{tabular}{ccccccccccccccccccccccc}
\toprule
\multirow{2}{*}{Method}  & \multirow{2}{*}{Times $\downarrow$} & & \multicolumn{3}{c}{110\_13051\_23361} & & \multicolumn{3}{c}{415\_57112\_110099} & & \multicolumn{3}{c}{106\_12648\_23157} & & \multicolumn{3}{c}{245\_26182\_52130} & & \multicolumn{3}{c}{34\_1403\_4393} \\ 
\cline{4-6} \cline{8-10} \cline{12-14} \cline{16-18} \cline{20-22} &  &  & PSNR $\uparrow$  & SSIM $\uparrow$  & LPIPS $\downarrow$   &  & PSNR  & SSIM  & LPIPS &  & PSNR & SSIM  & LPIPS  &  & PSNR   & SSIM & LPIPS  &  & PSNR   & SSIM & LPIPS \\ \hline
Nope-NeRF~\cite{bian2023nope} & $\thicksim$30 h & & 26.86 & 0.73 & 0.47 & & 24.78 & 0.64 & 0.55 & & 20.41 & 0.46 & 0.58 & & 25.05 & 0.80 & 0.49 & & 28.62 & 0.80 & 0.35\\
Ours & \textbf{$\thicksim$2 h} & & \textbf{29.69} & \textbf{0.89} & \textbf{0.29} & & \textbf{26.21} & \textbf{0.73} & \textbf{0.32} & & \textbf{22.14}	& \textbf{0.64} & \textbf{0.34} & & \textbf{27.24} & \textbf{0.85} & \textbf{0.30} & &  27.75 & \textbf{0.86} & \textbf{0.20} \\
\bottomrule
\end{tabular}
\vspace{-2mm}
\caption{\textbf{Novel view synthesis results on CO3D V2}. Each baseline method is trained with its public code under the original settings and evaluated with the same evaluation protocol. The best results are highlighted in bold.}

\vspace{-0.05in}
\label{table:nvs_co3d}
\end{table*}
\begin{table*}[ht]
\centering
\setlength{\tabcolsep}{3pt}
\footnotesize
\begin{tabular}{ccccccccccccccccccccccc}
\toprule
\multirow{2}{*}{Method}  & \multirow{2}{*}{Times $\downarrow$} & & \multicolumn{3}{c}{110\_13051\_23361} & & \multicolumn{3}{c}{415\_57112\_110099 } & & \multicolumn{3}{c}{106\_12648\_23157} & & \multicolumn{3}{c}{245\_26182\_52130} & & \multicolumn{3}{c}{34\_1403\_4393} \\ 
\cline{4-6} \cline{8-10} \cline{12-14} \cline{16-18} \cline{20-22} &  &  & $\text{RPE}_t \downarrow$ & $\text{RPE}_r \downarrow$ & ATE$ \downarrow$ &  & $\text{RPE}_t $ & $\text{RPE}_r $ & ATE &  &  $\text{RPE}_t $ & $\text{RPE}_r $ & ATE &  & $\text{RPE}_t$ & $\text{RPE}_r$ & ATE   &  & $\text{RPE}_t$ & $\text{RPE}_r$ & ATE   \\ \hline
Nope-NeRF~\cite{bian2023nope} & $\thicksim$30 h & & 0.400 & 1.966 & 0.046 & & 0.326	& 1.919	& 0.054 & & 0.387 & 1.312 & 0.049 & & 0.587 & 1.867 & 0.038 & & 0.591 & 1.313 & 0.053 \\
Ours & \textbf{$\thicksim$2 h} & & \textbf{0.140} & \textbf{0.401} & \textbf{0.021} &  &  \textbf{0.110} & \textbf{0.424} & \textbf{0.014} & & \textbf{0.094} & \textbf{0.360} & \textbf{0.008} & & \textbf{0.239}	& \textbf{0.472} & \textbf{0.017} & & \textbf{0.505} & \textbf{0.211} & \textbf{0.009}  \\
\bottomrule
\end{tabular}

\vspace{-2mm}
\caption{\textbf{Pose accuracy on CO3D V2}. Note that the camera poses provided by CO3D as the ``ground truth". The unit of $\text{RPE}_r$ is in degrees, ATE is in the ground truth scale and $\text{RPE}_t$ is scaled by 100. The best results are highlighted in bold.}
\vspace{-0.05in}
\label{table:pose_co3d}
\end{table*}
\begin{table}[tp]
\setlength{\tabcolsep}{2pt}
\footnotesize
\begin{tabular}{ccccccccccccccccccccccc}
\toprule
\multirow{2}{*}{scenes} & & \multicolumn{5}{c}{w.o. growing} & & \multicolumn{5}{c}{Ours} \\ 
\cline{3-4} \cline{6-7} \cline{9-10} \cline{12-13} 
 &  & PSNR  & SSIM & & $\text{RPE}_t$  &$\text{RPE}_r$ & & PSNR & SSIM & & $\text{RPE}_t$ & $\text{RPE}_r$ \\ \hline
Church & & 22.01 & 0.72 & & 0.044 & 0.122 & & 30.23	& 0.93 & & 0.008 & 0.018 \\ 
Barn & & 25.20 & 0.85 & & 0.152	& 0.232 & & 31.23 & 0.90 & & 0.034 & 0.034 \\ 
Museum & & 20.95 & 0.70 & & 0.079 & 0.212 & & 29.91	& 0.91 & & 0.052 & 0.215 \\ 
Family & & 22.30 & 0.77 & & 0.065 & 0.028 & & 31.27	& 0.94 & & 0.022 & 0.024 \\
Horse & & 23.47	& 0.81 & & 0.147 & 0.066 & & 33.94 & 0.96 & & 0.112	& 0.057\\
Ballroom & & 23.36 & 0.79 & & 0.056	& 0.073 & & 32.47 & 0.96 & & 0.037 & 0.024 \\ 
Francis & & 22.20 & 0.69 & & 0.147 & 0.161 & & 32.72 & 0.91 & & 0.029 & 0.154 \\ 
Ignatius & & 21.05 & 0.67 & & 0.24 & 0.058 & & 28.43 & 0.90 & & 0.033 & 0.032\\ \hline
mean & & 22.57 & 0.75 & & 0.116	& 0.119 & & \textbf{31.28} & \textbf{0.93} & & \textbf{0.041} & \textbf{0.069} \\
\bottomrule
\end{tabular}

\caption{\textbf{Ablation for Progressively Growing on Tanks and Temples.} Performance on both novel view synthesis and camera pose estimation. The best results are highlighted in bold.}

\vspace{-4mm}
\label{table:abs_grow}
\end{table}

\textbf{Camera Pose Estimation.} The learnt camera poses are post-processed by Procrustes analysis as in~\cite{lin2021barf, bian2023nope} and compared with the ground-truth poses of training views. The quantitative results of camera pose estimation are summarized in Table~\ref{table:pose}. Our approach achieves comparable performance with the current state-of-the-art results. 
We hypothesize that the relatively poorer performance in terms of RPE$_r$ and RPE$_t$ may be attributed to relying solely on photometric loss for relative pose estimation in a local region. In contrast, Nope-NeRF incorporates additional constraints on relative poses beyond photometric loss, including the chamfer distance between two point clouds. As indicated in~\cite{bian2023nope}, omitting the point cloud loss leads to a significant decrease in pose accuracy.

\begin{figure*}[tp]
    \centering
    \bgroup 
     \def\arraystretch{0.1} 
     \setlength\tabcolsep{0.2pt}
     \begin{tabular}{cc}
     \includegraphics[width=0.5\linewidth]{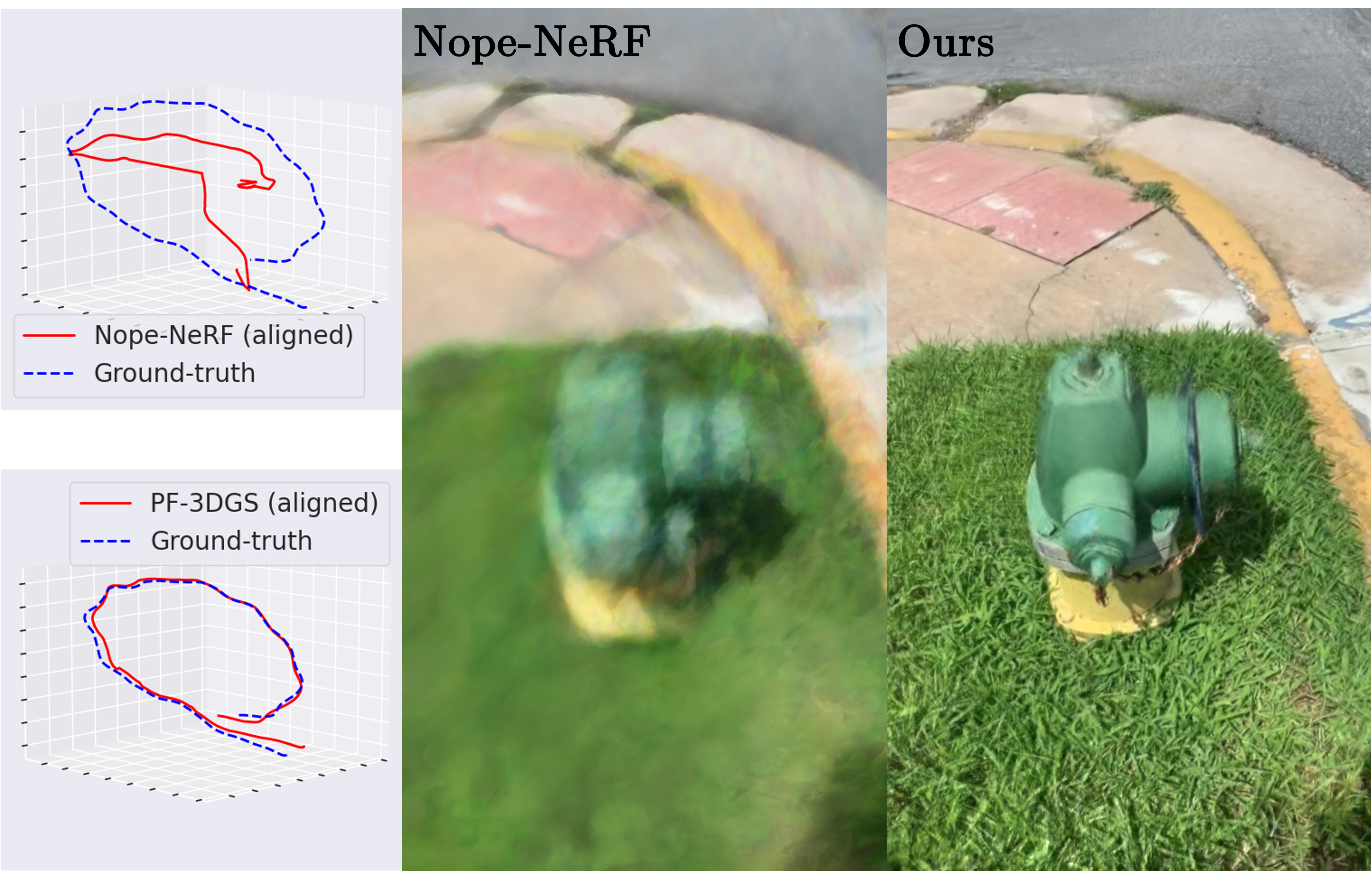} &
     \includegraphics[width=0.5\linewidth]{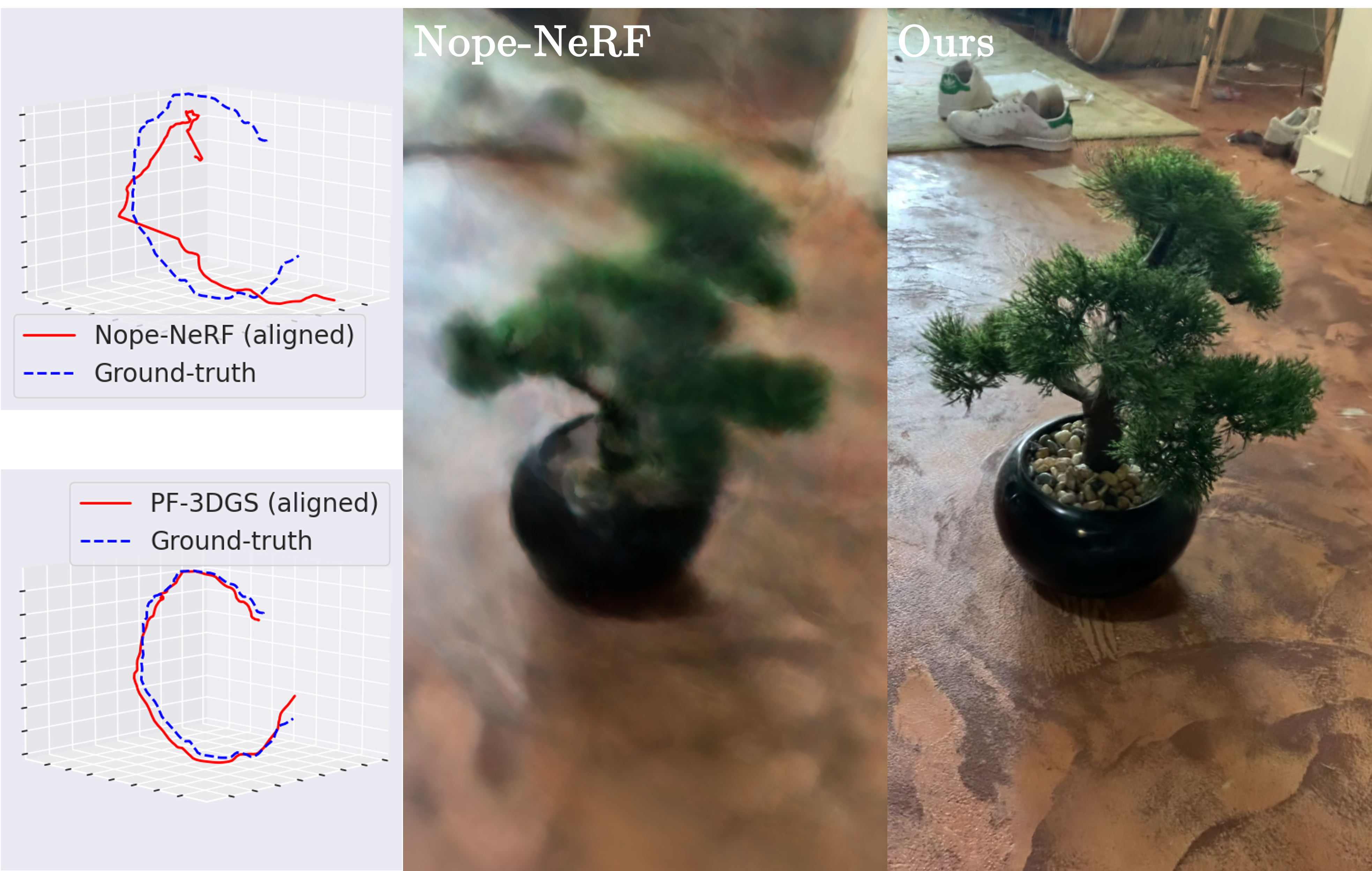}
     \end{tabular} \egroup

    \vspace{-2mm}
    \caption{\textbf{Qualitative comparison for novel view synthesis and camera pose estimation on CO3D V2.} Our approach estimates camera pose much more robust than Nope-NeRF, and thus generates higher quality rendering images. \textbf{Better viewed when zoomed in.}}
    \label{fig:co3d}
\end{figure*}

\vspace{-5mm}
\subsection{Results on Scenes with Large Camera Motions}
Given that the camera motion in scenes from the Tanks and Temples dataset is relatively small, we further demonstrate the robustness of our approach by validating it on the CO3D videos, which present more complex and challenging camera movements. We first evaluate the quality of synthesised images following the same evaluation procedure used in Tanks and Temples data. As demonstrated in Table~\ref{table:nvs_co3d}, for novel view synthesis, our approach also significantly outperforms Nope-NeRF which corroborates the conclusions drawn from experiments conducted on the Tanks and Temples dataset. More qualitative results are shown in Fig.~\ref{fig:co3d}.

In addition, we evaluated camera pose estimation on the CO3D V2 dataset with the provided ground-truth poses for reference. As detailed in Table~\ref{table:pose_co3d}, different from the on-par results on Tanks and Temples, our approach consistently surpasses Nope-NeRFe across all metrics by a large margin when testing on CO3D V2. This enhanced performance demonstrates the robustness and accuracy of our proposed method in estimating camera poses, especially in scenarios with complex camera movements.

\begin{table}[tp]
\setlength{\tabcolsep}{2pt}
\footnotesize
\begin{tabular}{ccccccccccccccccccccccc}
\toprule
\multirow{2}{*}{scenes} & & \multicolumn{5}{c}{w. depth} & & \multicolumn{5}{c}{Ours} \\ 
\cline{3-4} \cline{6-7} \cline{9-10} \cline{12-13} 
 &  & PSNR  & SSIM & & $\text{RPE}_t$  &$\text{RPE}_r$ & & PSNR & SSIM & & $\text{RPE}_t$ & $\text{RPE}_r$ \\ \hline
Church & & 28.93 & 0.91 & & 0.008 & 0.017 & & 30.23 & 0.93 & & 0.008 & 0.018 \\ 
Barn & & 28.70 & 0.87 & & 0.029 & 0.033 & & 31.23 & 0.90 & & 0.034 & 0.034 \\ 
Museum & & 26.92 & 0.83 & & 0.049 & 0.216 & & 29.91	& 0.91 & & 0.052 & 0.215 \\ 
Family & & 29.05 & 0.94 & & 0.021 & 0.024 & & 31.27 & 0.94 & & 0.022 & 0.024 \\
Horse & & 30.86	& 0.94 & & 0.108 & 0.054 & & 33.94 & 0.96 & & 0.112	& 0.057\\
Ballroom & & 30.38 & 0.94 & & 0.038	& 0.018 & & 32.47 & 0.96 & & 0.037 & 0.024 \\ 
Francis & & 29.97 & 0.88 & & 0.029 & 0.154 & & 32.72 & 0.91 & & 0.029 & 0.154 \\ 
Ignatius & & 26.69 & 0.87 & & 0.032	& 0.033 & & 28.43 & 0.90 & & 0.033 & 0.032\\ \hline
mean & & 28.94 & 0.90 & & 0.039 & 0.069 & & \textbf{31.28} & \textbf{0.93} & & 0.041 & 0.069 \\
\bottomrule
\end{tabular}
\caption{\textbf{Ablation study of depth loss on Tanks and
Temples.} We report the performance on both novel view synthesis and camera pose estimation. The best results are highlighted in bold.}
\vspace{-0.08in}
\label{table:abs_depth}
\end{table}

\subsection{Ablation Study}
In this section, we analyse the effectiveness of different pipeline designs and components that have been added to our approach.

\textbf{Effectiveness of Progressively Growing.} We first validate the effectiveness of progressively growing by removing it from the optimization of the global 3DGS. 
In other words, we change the current one-stage pipeline to a two-stage process, where the camera pose estimation and 3D Gaussian splatting are learnt in two separate steps. We report the performance on novel view synthesis and camera poses estimation w./w.o progressively growing in Table~\ref{table:abs_grow}. We observe that the progressively growing is essential for enhancing both novel view synthesis and pose estimation. Without progressive growth, 3DGS is unable to utilize the continuity present in videos, which results in the unstable optimization of the global 3DGS model.

\textbf{RGB Loss vs Depth Loss.} Depth-related losses play a crucial role in some advanced pose-unknown approaches, such as Nope-NeRF~\cite{bian2023nope}. To evaluate the significance of depth-related loss,  we employed both RGB and depth loss as the objective function during the optimization. As listed in Table~\ref{table:abs_depth}, we observe that the depth loss is not as effective as used in NeRFs. The performance on novel view synthesis is even better when merely using the photometric loss. 

\textbf{Comparison with 3DGS with COLMAP poses.} We also compare the novel view synthesis quality of our proposed framework against the original 3DGS~\cite{kerbl20233d}, which was trained using COLMAP-derived poses on the Tanks and Temples dataset. As indicated in Table~\ref{table:abs_colmap}, our joint optimization framework achieves performance comparable to the 3DGS model trained with COLMAP-assisted poses.

\begin{table}[tp]
\centering
\setlength{\tabcolsep}{3pt}
\small
\begin{tabular}{cccccccccc}
\toprule
\multirow{2}{*}{scenes} & & \multicolumn{3}{c}{Ours} & & \multicolumn{3}{c}{COLMAP + 3DGS} \\ 
\cline{3-5} \cline{7-9}
 &  & PSNR  & SSIM & LPIPS & & PSNR & SSIM & LPIPS \\ \hline
Church & & 30.23 & 0.93 & 0.11 & & 29.93 & 0.93 & 0.09 \\
Barn & & 31.23	& 0.90 & 0.10 & & 31.08 & 0.95 & 0.07 \\
Museum & & 29.91 & 0.91 & 0.11 & & 34.47 & 0.96 & 0.05  \\ 
Family & & 31.27 & 0.94 & 0.07 & & 27.93 & 0.92 & 0.11\\
Horse & & 33.94	& 0.96	& 0.05 & & 20.91 & 0.77 & 0.23\\
Ballroom & & 32.47 & 0.96 & 0.07 & & 34.48 & 0.96 & 0.04 \\ 
Francis & & 32.72 & 0.91 & 0.14 & & 32.64 & 0.92 & 0.15 \\ 
Ignatius & & 28.43 & 0.90 & 0.09 & & 30.20 & 0.93 & 0.08 \\ \hline
mean & & \textbf{31.28} & \textbf{0.93} & \textbf{0.09} & & 30.20 & 0.92 & 0.10\\
\bottomrule
\end{tabular}
\vspace{-2mm}
\caption{\textbf{Comparison to 3DGS trained with COLMAP poses}. We
report the performance of novel view synthesis using ours and vanilla 3DGS. The best results are highlighted in bold}
\vspace{-3mm}
\label{table:abs_colmap}
\end{table}

\section{Conclusion}
\label{sec:conclusion}
In this work, we present CF-3DGS, an end-to-end framework for joint camera pose estimation and novel view synthesis from a sequence of images. We demonstrate that previous works either have difficulty handling large camera motions or require extremely long training durations. Diverging from the implicit representation of NeRFs, our approach utilizes explicit point clouds to represent scenes. Leveraging the capabilities of 3DGS and the continuity inherent in video streams, our method sequentially processes input frames, progressively expanding the 3D Gaussians to reconstruct the entire scene. We show the effectiveness and robustness of our approach on challenging scenes like 360$^{\circ}$ videos. Thanks to the advantages of Gaussian splatting, our approach achieves rapid training and inference speeds. 

\paragraph{Limitations.} Our proposed method optimizes camera pose and 3DGS jointly in a sequential manner, thereby restricting its application primarily to video streams or ordered image collections. Exploring extensions of our work to accommodate unordered image collections represents an intriguing direction for future research.

\vspace{-3mm}
{\paragraph{Acknowledgements.}  This paper was done as a part-time internship project at NVIDIA LPR. Besides, it was supported, in part, by NSF CAREER Award IIS-2240014, Amazon Research Award, Intel Rising Star Faculty Award, and Qualcomm Innovation Fellowship.}

\newpage
{\small
\bibliographystyle{ieeenat_fullname}
\bibliography{11_references}
}

\ifarxiv \clearpage \appendix \maketitlesupplementary
\appendix
\section{Implementation Details}

\subsection{Dataset}
We select sequences containing dramatic camera motions Tanks and Tamples~\cite{Knapitsch2017} and CO3D-V2~\cite{reizenstein2021common} for training and evaluation. The details of each sequence are listed in Table~\ref{table:data}, where \textit{Max rotation} denotes the maximum relative rotation angle between any two frames in a sequence. 
The sampled images are further split into training and test sets. Starting from the 5\textit{th} image, we sample every 8\textit{th} image in a sequence as a test image. However, this leads to a change in the sampling rate in the temporal domain among training images. In order to study the effect of the sampling rate changes, we follow the experiment setting proposed by~\cite{bian2023nope}. Specifically, for scene \textit{Family} in Tanks and Temples~\cite{Knapitsch2017}, we sample every other image as test images, \ie, training on images with odd frame ids and testing on images with even frame ids. For CO3D-V2~\cite{reizenstein2021common}, we randomly select 10 scenes from 6 categories, \eg, apple, bench, hydrant, plant, skateboard, and teddybear. The selected sequence IDs are also shown in Table~\ref{table:data} (bottom part). Compared to Tanks and Temples, most scenes achieve the \textit{Max rotation} of $180^{\circ}$ indicating more dramatic and larger camera motions than Tanks and Temples.

\begin{table}[ht]
\resizebox{\linewidth}{!}{%
\begin{tabular}{lccccc}
\hline
& Scenes   & Type    & Seq. length & Frame rate & Max. rotation (deg) \\ \hline
\multirow{8}{*}{\rotatebox[origin=c]{90}{Tanks and Temples}} 
& Church   & indoor  & 400          & 30            & 37.3     \\
& Barn     & outdoor & 150          & 10            & 47.5     \\
& Museum   & indoor  & 100          & 10            & 76.2     \\
& Family   & outdoor & 200          & 30            & 35.4     \\
& Horse    & outdoor & 120          & 20            & 39.0     \\
& Ballroom & indoor  & 150          & 20            & 30.3     \\
& Francis  & outdoor & 150          & 10            & 47.5     \\
& Ignatius & outdoor & 120          & 20            & 26.0     \\ \hline
\multirow{10}{*}{\rotatebox[origin=c]{90}{CO3D-V2}}
& 34\_1403\_4393       & indoor    & 202    & 30    & 180.0     \\
& 106\_12648\_23157    & outdoor   & 202    & 30    & 180.0     \\
& 110\_13051\_23361    & indoor    & 202    & 30    & 71.6     \\
& 219\_23121\_48537    & indoor    & 202    & 30    & 180.0     \\
& 245\_26182\_52130    & indoor    & 202    & 30    & 180.0     \\ 
& 247\_26441\_50907    & indoor    & 202    & 30    & 180.0     \\ 
& 407\_54965\_106262   & indoor    & 202    & 30    & 180.0     \\
& 415\_57112\_110099   & outdoor   & 202    & 30    & 180.0     \\
& 415\_57121\_110109   & outdoor   & 202    & 30    & 180.0     \\
& 429\_60388\_117059   & outdoor   & 202    & 30    & 180.0     \\
\hline

\end{tabular}
}
\caption{\textbf{Details of selected sequences.} We downsample several videos to a lower frame rate. FPS denotes frame per second. \textit{Max rotation} denotes the maximum relative rotation angle between any two frames in a sequence. Our method can handle dramatic camera motion (large maximum rotation angle).}
\label{table:data}
\end{table}

\subsection{Training Details.}
\begin{algorithm}[!h]
    \caption{Local 3DGS Optimization}
    \label{alg:local}
    \begin{algorithmic}
        \State $\{I_t, I_{t+1}\} \gets$ Two nearby images
        \State $\text{DPT} \gets $ Monocular Depth Estimation Model
        \State $D_t \gets \text{DPT}(I_t)$ 
        \State $G_t \gets$ InitGauss($I_t$, $D_t$) \Comment{Init Local 3DGS}
        \State $T_t \gets $ Identity $\mathbb{I}$  \Comment{Init Pose}
        \While{not converged} 
            \State $\hat{I}_{t} \gets $ Rasterize($G_t$)
            \State $L \gets \text{Loss}(I_t, \hat{I}_t) $
            \State ${G_t} \gets$ Adam($\nabla L$) \Comment{Update Local 3DGS}
        \EndWhile
        \While{not converged} 
            \State $\hat{I}_{t+1} \gets $ Rasterize($T_t \odot G_t$)
            \State $L \gets \text{Loss}(I_{t+1}, \hat{I}_{t+1}) $
            \State ${T_t}^* \gets$ Adam($\nabla L$) \Comment{Update Pose}
        \EndWhile
        \State $T_t \gets \prod_{i=1}^{t}T_{i}$ \Comment{Output Pose}
        
        
        
        
        

    \end{algorithmic}
\end{algorithm}

\noindent\textbf{Local 3DGS.} During the training of local 3DGS, we first obtain the monocular depth map of the input image by pre-trained monocular depth estimator, \ie, DPT~\cite{ranftl2021vision}, ZeoDepth~\cite{bhat2023zoedepth}. Then, the depth map is lifted up with the given camera intrinsic. As the high-resolution input images could lead to a huge amount of point clouds, we downsample the point cloud first before fitting it by 3DGS. Then, the downsampled point cloud is used to initialize the local 3DGS and is further optimized on the input view via photometric loss for 500 iterations. To obtain the transformation of the 3D Gaussian between two views, we freeze the pre-trained local 3DGS including all attributes (\ie, position, SH coefficient, opacity, scale, and rotation), and learn the pose parameter of a quaternion vector a translation vector by the photometric loss between the target view and the rendering image. In detail, the freeze local 3D Gaussian is first transformed into the target view coordinate by the learnable pose parameter and then rendered into the target view by the gaussian splatting. The optimization of the camera pose learning process takes 300 steps. The optimization algorithm of local 3DGS is summarized in Algorithm~\ref{alg:local}

\noindent\textbf{Global 3DGS.} The optimization process of the global 3DGS 
starts and initializes from the first frame and its monocular depth estimation. Subsequently, camera poses are estimated in a sequential manner using the local 3DGS, as described in Algorithm~\ref{alg:local}. Concurrently, the global 3DGS is updated with all the observed images to date (\ie, from the first to the current image), in tandem with the camera pose estimation. As each new frame is introduced, the global 3DGS progressively grows and expands through a densification process. 

\subsection{Evaluation Metrics}
\noindent\textbf{Novel View Synthesis.}
We use Peak Signal-to-Noise Ratio (PSNR), Structural Similarity Index Measure (SSIM)~\cite{wang2004image}, and Learned Perceptual Image Patch Similarity (LPIPS)~\cite{zhang2018unreasonable} to measure the novel view synthesis quality. For LPIPS, we use a VGG architecture~\cite{simonyan2014very}.

\noindent\textbf{Pose Accuracy.}
To evaluate pose accuracy, we employ standard visual odometry metrics, including Absolute Trajectory Error (ATE) and Relative Pose Error (RPE). ATE quantifies the discrepancy between estimated camera positions and their ground truth counterparts. RPE, on the other hand, assesses the errors in relative poses between image pairs. This includes both relative rotation error ($\text{RPE}_r$) and relative translation error ($\text{RPE}_t$). 

\section{Additional Experiments}
The subsequent sections present further quantitative and qualitative results of novel view synthesis and camera pose estimation, conducted on both the Tanks and Temples and CO3D-V2 datasets.

\subsection{Camera Pose Estimation}
\noindent\textbf{Additional results on CO3D-V2.} 
We conduct experiments on 5 additional scenes of the CO3D-V2 dataset for the task of camera pose estimation. The results are reported in Table~\ref{table:supp_co3d_pose}. We show better performances than Nope-NeRF~\cite{bian2023nope} in both pose accuracy and synthesis quality.
\begin{table}[tp]
\centering
\setlength{\tabcolsep}{3pt}
\footnotesize
\begin{tabular}{cccccccccc}
\toprule
\multirow{2}{*}{scenes} & & \multicolumn{3}{c}{Nope-NeRF} & & \multicolumn{3}{c}{Ours} \\ 
\cline{3-5} \cline{7-9}
 &  & $\text{RPE}_t$ & $\text{RPE}_r$ & ATE & & $\text{RPE}_t$ & $\text{RPE}_r$ & ATE \\ \hline
189\_20393\_38136 & & 0.444	& 2.84 & 0.034 & & 0.064 & 0.225 & 0.007 \\
247\_26441\_50907 & & 0.34 & 1.395 & 0.032 & & 0.395 & 0.477 & 0.007 \\
407\_54965\_106262 & & 0.553 & 4.685 & 0.057 & & 0.31 & 0.243 & 0.008 \\ 
429\_60388\_117059 & & 0.398 & 2.914 & 0.055 & & 0.134	& 0.542 & 0.018	 \\
46\_2587\_7531     & &  0.426 & 4.226 & 0.023 & & 0.095 & 0.447 & 0.009	 \\ \hline
mean & & 0.432 & 3.212 & 0.040 & & \textbf{0.200} & \textbf{0.387} & \textbf{0.010}	\\
\bottomrule
\end{tabular}
\vspace{-2mm}
\caption{\textbf{Camera Pose Estimation on CO3D V2.} 
The best results are highlighted in bold.}
\vspace{-4mm}
\label{table:supp_co3d_pose}
\end{table}

\noindent\textbf{Additional Visualization.}
Additional qualitative results for camera pose estimation on CO3D-V2 are presented in Fig.~\ref{fig:supp_co3d_cam}, following the evaluation procedure outlined in the main paper. In scenarios involving large camera motions, our approach significantly outperforms Nope-NeRF.
\begin{figure*}[tp]
    \centering
    \includegraphics[width=1.0\linewidth]{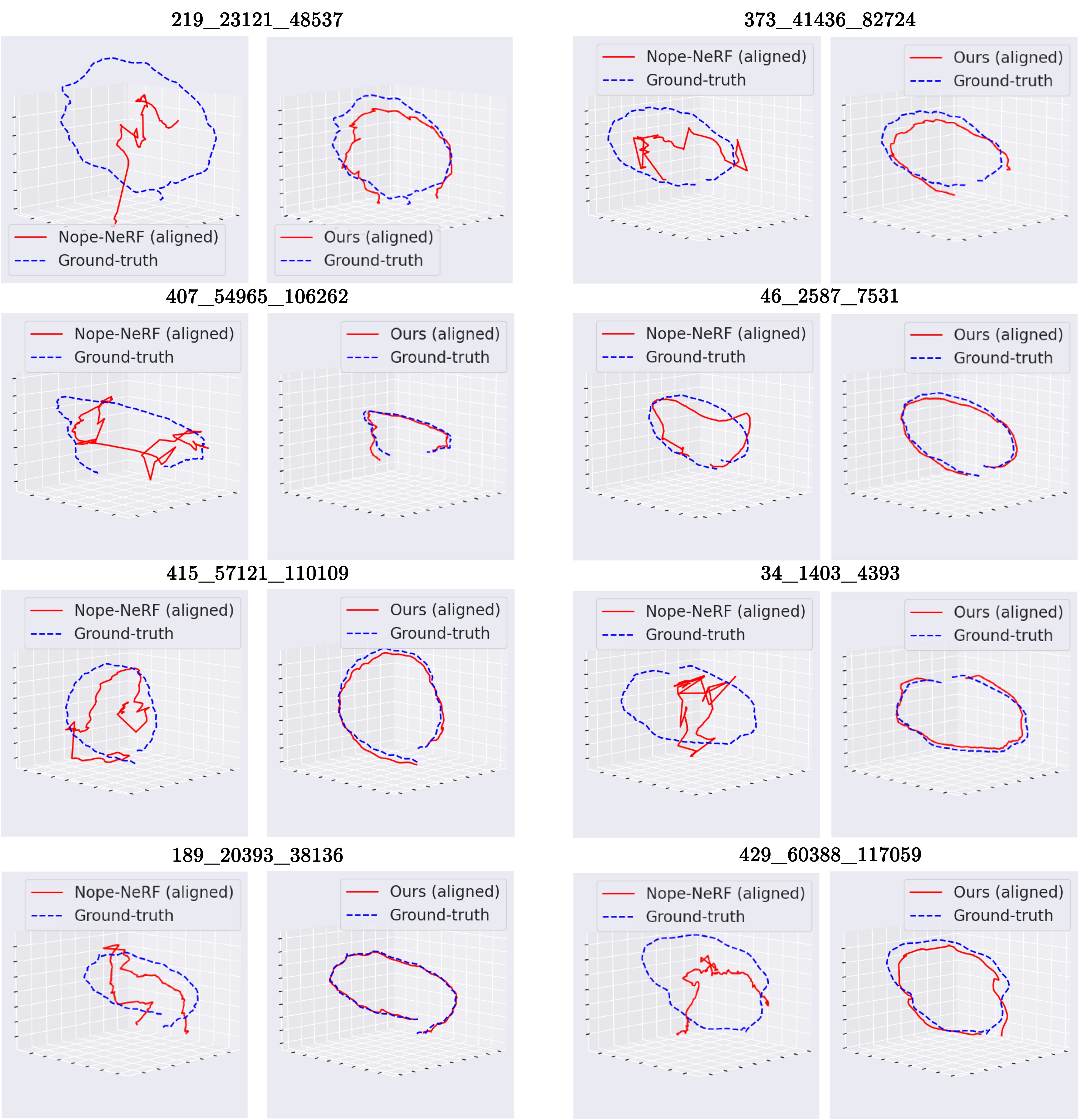}
    \caption{\textbf{Qualitative comparison for Camera Pose Estimation on CO3D-V2.} The ground-truth trajectory and the estimated one are shown in blue and red, respectively. }
    \label{fig:supp_co3d_cam}
\end{figure*}

\subsection{Novel View Synthesis.} 
\noindent\textbf{Render Novel Views.}
As mentioned in the main paper, we minimize the photometric error of the synthesized images while freezing the 3DGS model to obtain the testing camera poses. Because the test views are sampled from videos that are close to the training views, these good results may be obtained due to overfitting to the training images. Therefore, we conduct an additional qualitative evaluation on more novel views. Specifically, we fit a bezier curve from the estimated training poses and sample interpolated poses for each method to render novel view videos. The rendered images are shown in Fig.~\ref{fig:supp_tanks_nvs_part1} and Fig.~\ref{fig:supp_tanks_nvs_part2}. Compared to Nope-NeRF~\cite{bian2023nope}, our approach renders photo-realistic images with more details (please check the highlighted regions). 

\begin{table}[tp]
\centering
\setlength{\tabcolsep}{3pt}
\footnotesize
\begin{tabular}{cccccccccc}
\toprule
\multirow{2}{*}{scenes} & & \multicolumn{3}{c}{Nope-NeRF} & & \multicolumn{3}{c}{Ours} \\ 
\cline{3-5} \cline{7-9}
 &  & PSNR  & SSIM & LPIPS & & PSNR & SSIM & LPIPS \\ \hline
189\_20393\_38136 & & 29.37 & 0.85 & 0.54 & & 32.41 & 0.92 & 0.26 \\
247\_26441\_50907 & & 23.49 & 0.73 & 0.54 & & 23.88	& 0.75 & 0.36 \\
407\_54965\_106262 & & 25.53 & 0.83 & 0.58 & & 27.80 & 0.84	& 0.35 \\ 
429\_60388\_117059 & & 22.19 & 0.62 & 0.56 & & 24.44 & 0.68 & 0.36 \\
46\_2587\_7531     & &  25.3 & 0.73 & 	0.46  & & 25.44 & 0.80 & 0.21 \\ \hline
mean & &  25.18	& 0.75 & 0.54 & &  \textbf{26.79} & \textbf{0.80} & \textbf{0.31} \\
\bottomrule
\end{tabular}
\vspace{-2mm}
\caption{\textbf{Novel view synthesis results on CO3D V2.} 
The best results are highlighted in bold.}
\vspace{-4mm}
\label{table:supp_co3d_nvs}
\end{table}
\begin{figure*}[tp]
    \centering
    \includegraphics[width=0.95\linewidth]{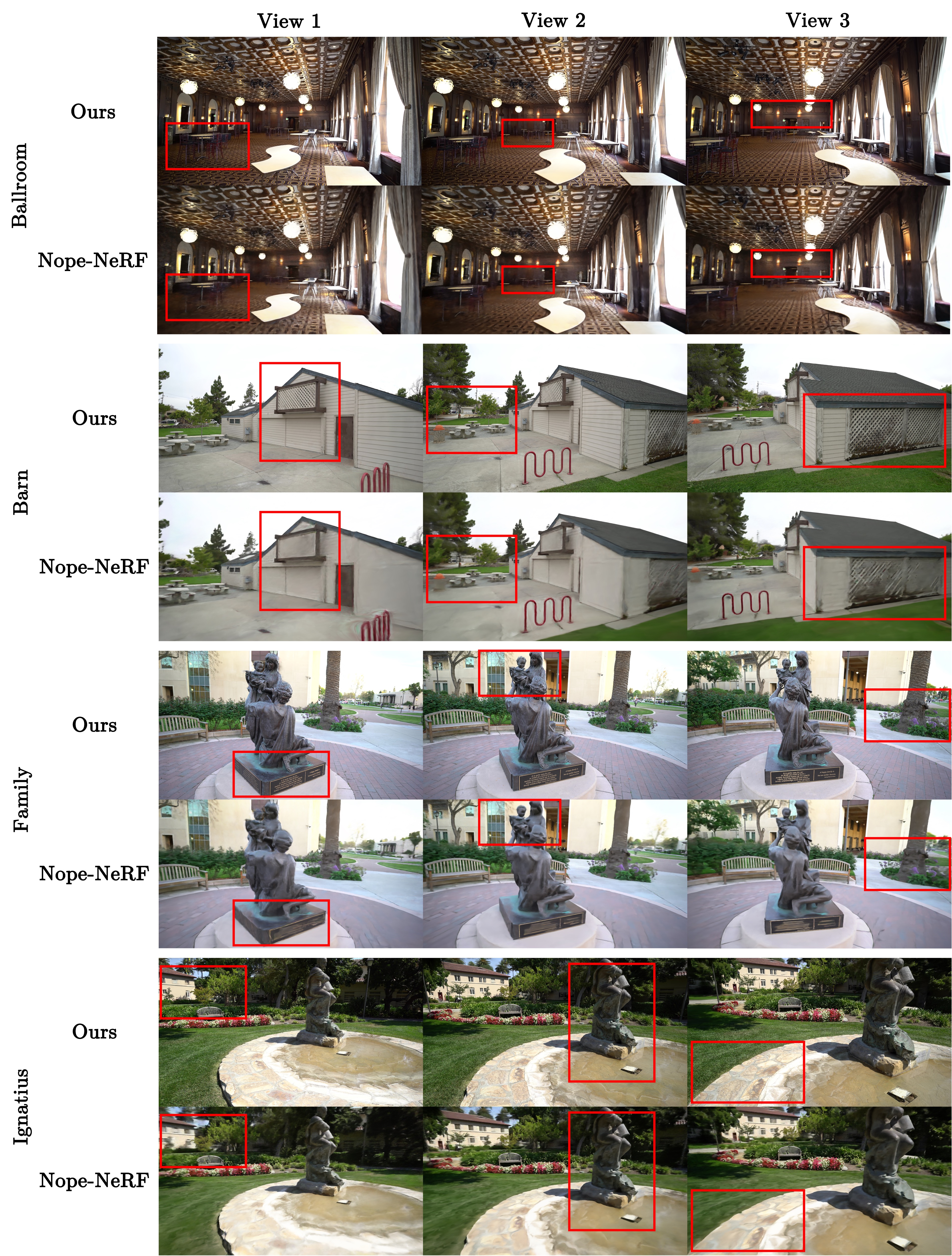}
    \caption{\textbf{Qualitative comparison for novel view synthesis on Tanks and Temples.} For each method, we fit the learned trajectory with a bezier curve and uniformly sample new viewpoints for rendering. \textbf{Better viewed when zoomed in.}}
    \label{fig:supp_tanks_nvs_part1}
\end{figure*}
\begin{figure*}[tp]
    \centering
    \includegraphics[width=1.0\linewidth]{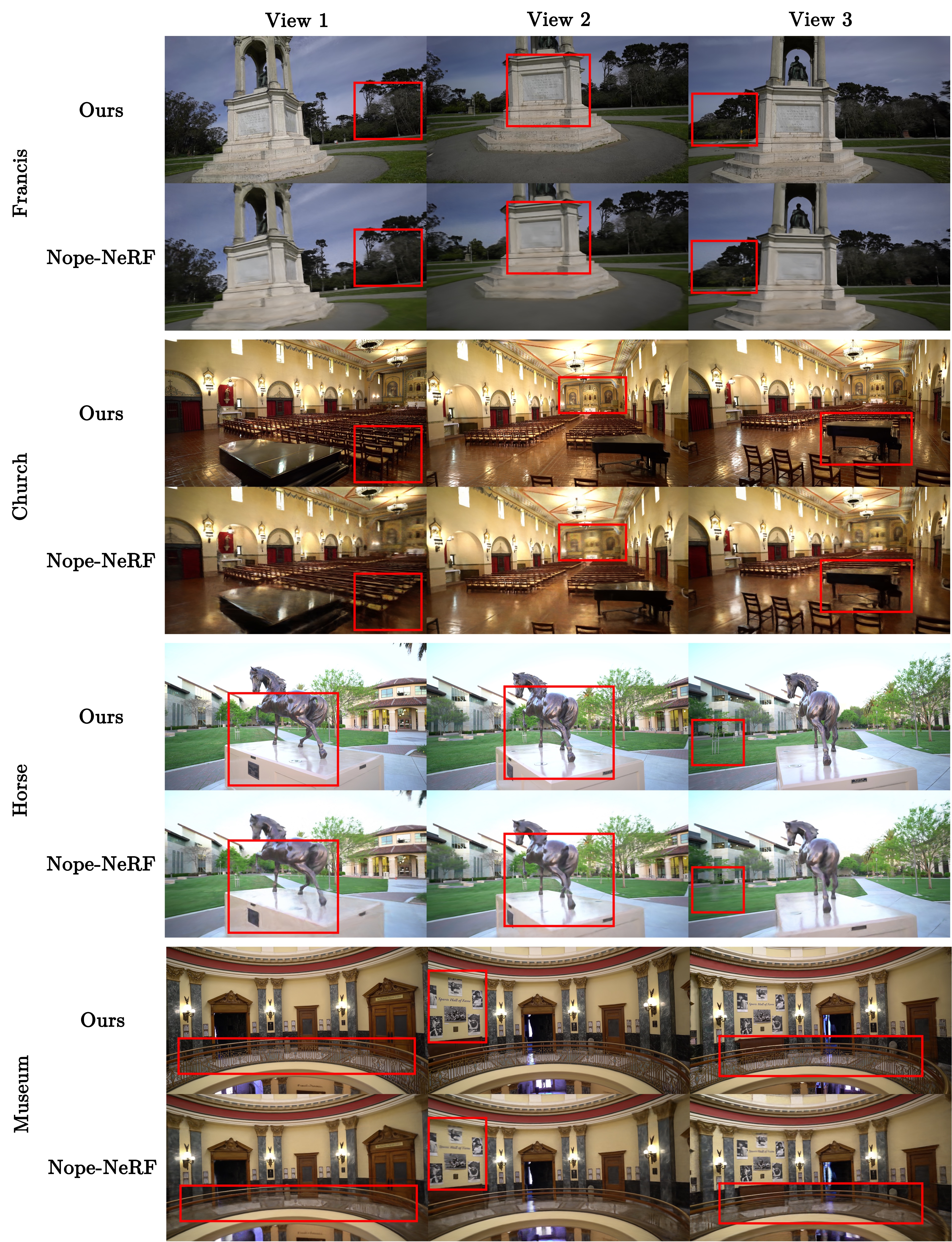}
    \caption{\textbf{Qualitative comparison for novel view synthesis on Tanks and Temples.} For each method, we fit the learned trajectory with a bezier curve and uniformly sample new viewpoints for rendering. \textbf{Better viewed when zoomed in.}}
    \label{fig:supp_tanks_nvs_part2}
\end{figure*}

\noindent\textbf{Unknown camera intrinsic}.
We also conduct experiments with heuristic camera intrinsic, where we set the FoV of all scenes to 79$^\circ$ and make the principle points to the image center. The quantitative results are listed in the following table. We find that by setting the camera intrinsic heuristically, the performance on novel view synthesis (NVS) and camera pose estimation slightly degenerates which is reasonable as the intrinsic parameters are also important and could be further optimized along with the camera extrinsic parameters.

\begin{table}[h]
\setlength{\tabcolsep}{3pt}
\footnotesize
\begin{tabular}{ccccccccccccccccccccccc}
\toprule

Method &  & PSNR  & SSIM & LPIPS  & & $\text{RPE}_t$ & $\text{RPE}_r$ & ATE \\ \hline
Heuristic Intrinsic & & 30.90 & 0.92 & 0.09 & & 0.044 & 0.072 & 0.004 \\ 
G.T. Intrinsic & & \textbf{31.28} & \textbf{0.93} & \textbf{0.09} & & \textbf{0.041} & \textbf{0.069} & \textbf{0.004}\\
\bottomrule
\end{tabular}
\vspace{-2mm}
\caption{{Ablation study of camera intrinsic on Tanks and
Temples.}}
\vspace{-2mm}
\label{table:rebuttal_cam_nvs}
\end{table}

\noindent\textbf{Different monocular depth estimator}.
We conduct ablation studies on different monocular depth estimation algorithms in the following table. We notice that more accurate monocular depth estimation results could always lead to better performance.
\begin{table}[h]
\setlength{\tabcolsep}{2pt}
\footnotesize

\begin{tabular}{ccccccccccccccccccccccc}
\toprule
\multirow{2}{*}{scenes} & & \multicolumn{5}{c}{ZeoDepth} & & \multicolumn{5}{c}{DepthAnything} \\ 
\cline{3-4} \cline{6-7} \cline{9-10} \cline{12-13}
 &  & PSNR  & SSIM & & $\text{RPE}_t$ &$\text{RPE}_r$  & & PSNR  & SSIM & & $\text{RPE}_t$ &$\text{RPE}_r$ \\ \hline
Church & & 30.49 & 0.93 & & 0.012 & 0.033 & & 30.66 & 0.93 & & 0.012 & 0.029 \\ 
Barn & & 28.34 & 0.86 & & 0.039	& 0.057 & & 30.54 & 0.88 & & 0.034 & 0.113\\ 
Museum & & 30.40 & 0.91 & & 0.052 & 0.158 & & 30.92	& 0.92 & & 0.043 & 0.130 \\ 
Family & & 28.79 & 0.91 & & 0.093 & 0.037 & & 32.54 & 0.95 & & 0.037 & 0.069 \\
Horse & & 33.32	& 0.95 & & 0.101 & 0.035 & & 33.96 & 0.96 & & 0.108	& 0.075 \\
Ballroom & & 32.86 & 0.96 & & 0.021	& 0.032 & & 32.54 & 0.96 & & 0.022 & 0.030 \\ 
Francis & & 31.05 & 0.89 & & 0.057 & 0.086 & & 32.73 & 0.91 & & 0.027 & 0.126 \\ 
Ignatius & & 22.75 & 0.75 & & 0.172	& 0.083 & & 28.89 & 0.89 & & 0.043 & 0.075\\ \hline
mean & & 29.75 & 0.90 & & 0.068 & 0.065 & & \textbf{31.60} & \textbf{0.93} & & \textbf{0.041} & 0.081\\
\bottomrule
\end{tabular}
\label{table:rebuttal_depth_nvs}
\vspace{-3mm}
\caption{{Ablation study of different depth estimators on Tanks and Temples.}}
\end{table}

\noindent\textbf{Additional results on CO3D-V2.} We conduct experiments on 5 additional scenes of the CO3D-V2 dataset and the novel view synthesis results are summarized in Table~\ref{table:supp_co3d_nvs}.

\noindent\textbf{Additional Visualization.}
We present additional qualitative results for novel view synthesis on Tanks and Temples and CO3D-V2 in Fig.~\ref{fig:supp_tanks_compare} and Fig.~\ref{fig:supp_co3d_nvs_part2} following the same evaluation procedure described in the main paper.

\begin{figure*}[tp]
    \centering
    \includegraphics[width=1.0\linewidth]{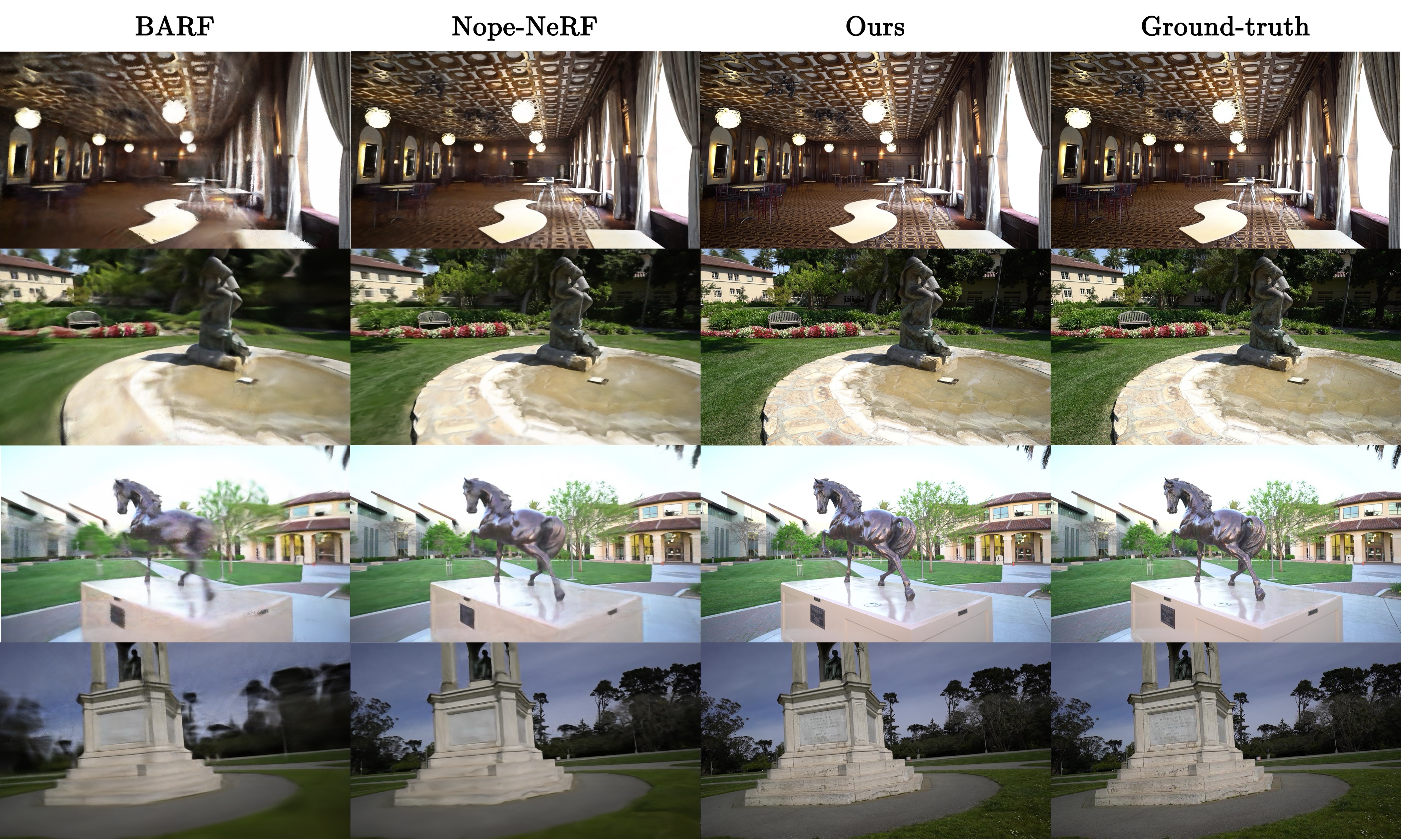}
    \vspace{-5mm}
    \caption{\textbf{Qualitative comparison for novel view synthesis on Tanks and Temples.} Our approach produces more realistic rendering results than other baselines. \textbf{Better viewed when zoomed in.}}
    \label{fig:supp_tanks_compare}
\end{figure*}
\begin{figure*}[tp]
    \centering
    \includegraphics[width=0.95\linewidth]{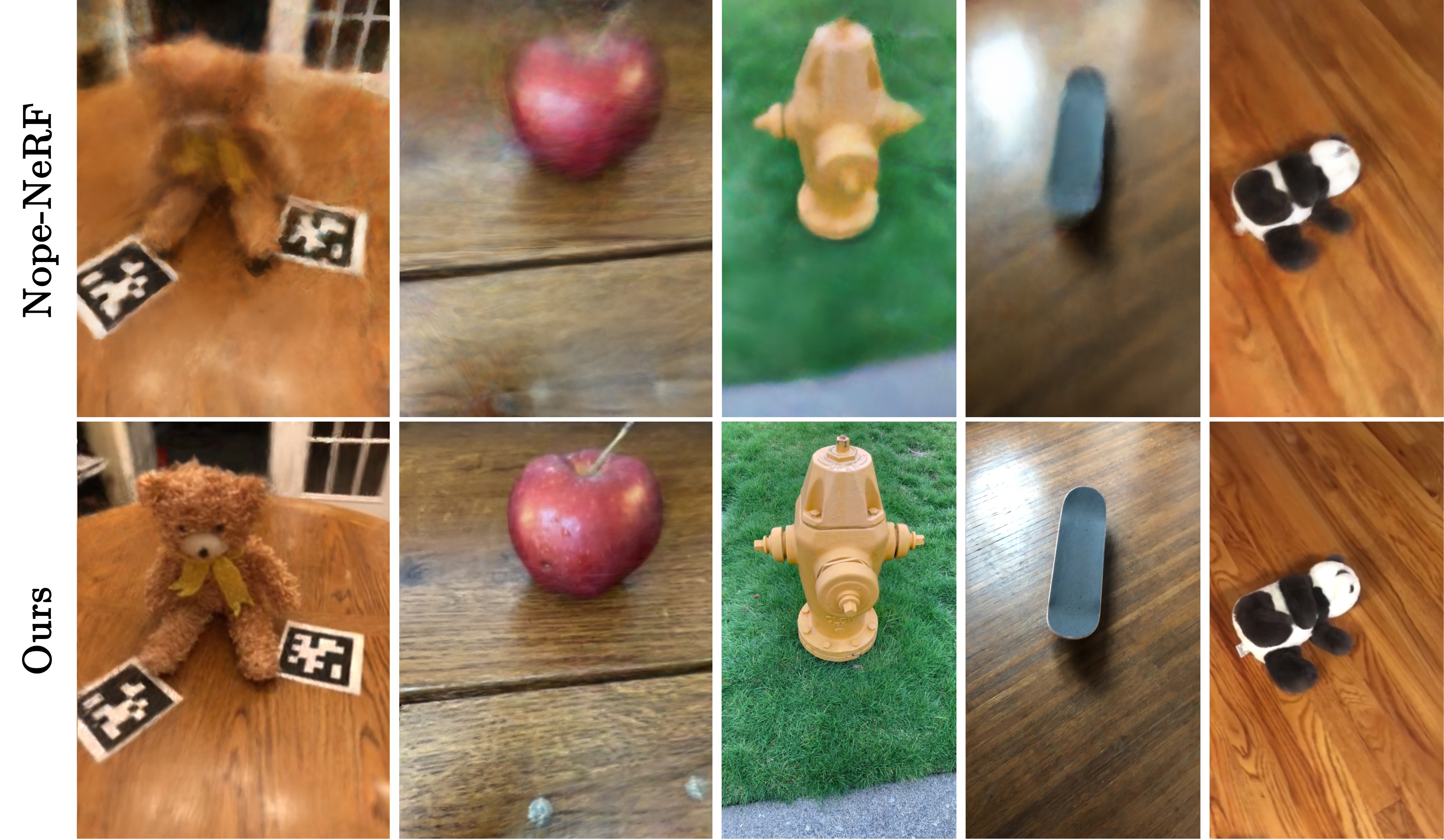}
    \vspace{-2mm}
    \caption{\textbf{Qualitative comparison for novel view synthesis on CO3D-V2.}Our approach produces more realistic rendering results than other baselines. Better viewed when zoomed in.}
    \label{fig:supp_co3d_nvs_part2}
\end{figure*}


 \fi

\end{document}